\documentclass{article}
\usepackage[preprint]{neurips_2026}
\usepackage{graphicx} %
\usepackage{enumitem}
\setlist[itemize]{topsep=2pt}
\usepackage{booktabs}
\usepackage{array}
\usepackage{xcolor}
\usepackage{colortbl}
\usepackage{pifont}
\usepackage{url}
\usepackage{pgfplots}
\usepackage{wrapfig}
\pgfplotsset{compat=1.18}
\usepgfplotslibrary{groupplots}
\usetikzlibrary{calc}
\usepackage{longtable}
\usepackage{amsmath}
\usepackage{amssymb}

\title{Visual Token Routing: Learning to keep tokens that matter}

\author{Pranjal Aggarwal \qquad Sean Welleck\\Carnegie Mellon University}
\date{November 2026}

\begin{document}

\maketitle

\begin{abstract}
Computer-use agents (CUAs) have recently achieved superhuman performance on several long-horizon benchmarks. However, their computational cost limits their use in long-horizon tasks and real-time domains such as video games. A key bottleneck is representing each screenshot as thousands of visual tokens, which accumulate into long contexts as agents reason over extended interaction histories. We observe that most of these visual tokens are either semantically irrelevant to the agent's next action or temporally redundant across screenshots. Building on this observation, we introduce Visual Token Routing (VTR), a joint training framework for learning sparse visual representations that preserve the information needed for action prediction. In VTR, we train the vision-language model together with a lightweight spatio-temporal router that examines each image region in the current and previous screenshots and decides whether its visual token enters the language model. We optimize the router using an adaptive loss to keep the number of visual tokens as small as possible while minimizing performance degradation. With just 2,000 training steps, VTR matches the performance of the full-resolution model on OSWorld, CUA-World, and video game action prediction, while using 2.5-5.8$\times$ fewer visual tokens. These savings reduce inference cost by up to 3.3$\times$ and speed up prefill by up to 2.9$\times$. VTR also outperforms lowering the image resolution and other baselines for dropping tokens, by up to 12.8 points on OSWorld. Overall, VTR enables computer-use agents to process only the visual tokens they need, allowing them to be used in long-horizon tasks and real-time domains at a fraction of the computational cost.\footnote{Code and models are available at \url{https://pranjal2041.github.io/vtr}.}
\end{abstract}

\section{Introduction}
\label{sec:introduction}

Computer-use agents (CUAs), which operate software by observing screenshots and issuing mouse and keyboard actions, have recently achieved superhuman performance on several long-horizon benchmarks~\citep{xie2024osworldbenchmarkingmultimodalagents,aggarwal2026gymanythingturnsoftwareagent,aggarwal2026cuaspeedrun}. However, their computational cost limits their use in long-horizon tasks and real-time domains such as video games~\citep{aggarwal2026cuaspeedrun,abhyankar2026osworldhumanbenchmarkingefficiencycomputeruse}. A key bottleneck is the representation of each screenshot as thousands of visual tokens, which accumulate into long contexts as agents reason over extended histories of interaction. For instance, Qwen3.5~\citep{qwen2026qwen35} encodes a single 1080p screenshot into about 2,000 visual tokens, so an agent that keeps its full history on a 100-step task has over 200,000 visual tokens in its context by the final step. In video games, agents observe the screen many times per second, and even at a reduced resolution of $640\times360$, one minute of play at 30 frames per second amounts to about 400,000 visual tokens. Furthermore, in real-time environments, speed is the primary bottleneck for frontier agents~\citep{aggarwal2026gymanythingrt}.

Existing methods reduce the number of visual tokens by keeping the tokens that receive the most attention within each image~\citep{chen2024fastv,yang2025visionzip}, or by dropping tokens that remain unchanged across video frames~\citep{shen2025longvu,bagrov2025evs}. However, these hand-designed rules keep a fixed fraction of tokens or apply a fixed threshold to every input, regardless of what the agent needs for its next action. Other work learns which visual tokens to keep jointly with the model for single images in general visual question answering~\citep{huang2025dynamicllava,ye2025atpllava}. However, computer-use agents face fundamentally new challenges, such as long histories of nearly identical screenshots, and actions that depend on small regions of the screen, like a cursor or a single spreadsheet cell.

\begin{figure}[t]
\centering
\includegraphics[width=\linewidth]{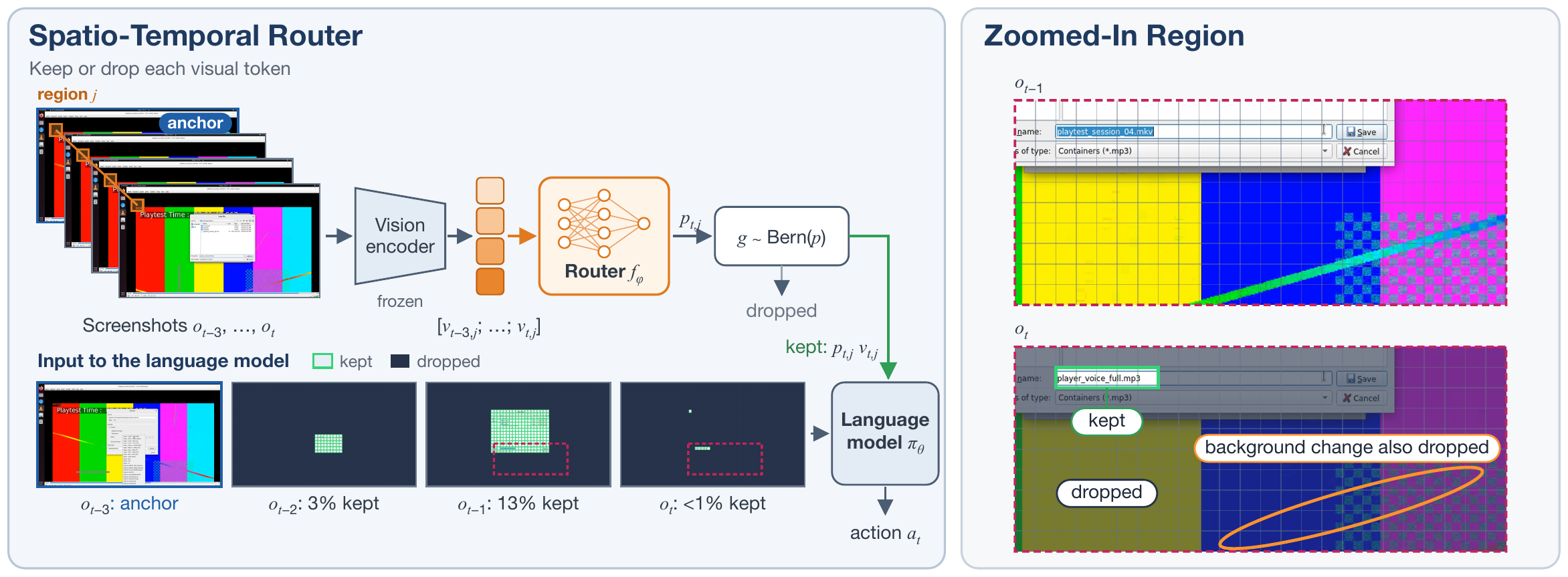}
\caption{\textbf{Overview of Visual Token Routing (VTR).} \textbf{Spatio-temporal router:} A lightweight router reads each image region across the current and previous screenshots and decides whether its visual token enters the language model. \textbf{Zoomed-in region:} While the agent types a file name into a save dialog in front of a playing video, the router keeps only the tokens of the typed file name. It drops the regions that did not change, and also drops the circled part of the video, which changed but is not relevant to the task.}
\label{fig:overview}
\end{figure}

We observe that most visual tokens in an agent's history are either semantically irrelevant to its next action or temporally redundant across screenshots. For instance, the taskbar and the empty parts of a document rarely affect the agent's next click, and consecutive game frames share most of their background. Building on this observation, we introduce Visual Token Routing (VTR), a joint training framework for learning sparse visual representations that preserve the information needed for action prediction (\S\ref{sec:method}; Figure~\ref{fig:overview}). In VTR, a lightweight spatio-temporal router processes each image region across the current and previous screenshots and outputs a decision to keep the visual token or not. Dropped tokens are removed from the input sequence, so the language model processes only the kept tokens, which retain their original positional encodings.

We train the router jointly with the language model to satisfy two objectives: (1) keeping as few visual tokens as possible, and (2) preserving the model's accuracy in predicting the agent's next action. However, training the router poses two challenges. First, dropping a token is a discrete decision that provides no gradient to the router, which we address by scaling each kept token by its keep probability, as in mixture-of-experts models~\citep{shazeer2017moe,fedus2022switch}. Second, unlike mixture-of-experts models, where each token is processed by a fixed number of experts, the router must also decide how many tokens to keep. We therefore measure how much the screenshots help the model predict each action, using reference models trained with and without them, and adaptively weight the two objectives so that the model keeps this benefit while dropping tokens. In doing so, the router learns how many visual tokens each input needs while preserving the model's performance, rather than relying on a fixed token budget.

We use VTR to train vision-language models on computer-use trajectories and video game play, and evaluate them on OSWorld~\citep{xie2024osworldbenchmarkingmultimodalagents}, CUA-World~\citep{aggarwal2026gymanythingturnsoftwareagent}, and video game action prediction. With just 2,000 training steps, VTR matches the performance of the full-resolution model in all three domains, while using 2.5-5.8$\times$ fewer visual tokens (Figure~\ref{fig:main-results}). These savings reduce inference cost by up to 3.3$\times$ and speed up prefill by up to 2.9$\times$ (Figure~\ref{fig:cost-speed}). In addition, VTR outperforms lowering the image resolution and hand-designed rules for dropping tokens, by up to 12.8 points on OSWorld. Furthermore, we find that learning visual token selection jointly with the model is crucial, improving performance by 13.2 points on OSWorld over the same router trained separately from the model.

In summary, our contributions are three-fold. First, we introduce Visual Token Routing (VTR), a joint training framework for learning sparse visual representations, in which a lightweight spatio-temporal router decides which visual tokens enter the language model. Second, we propose a dual objective that trains the router to keep as few visual tokens as possible while preserving action prediction, so that the router learns how many visual tokens each input needs without a fixed token budget. Third, through extensive evaluations, we show that VTR matches full-resolution performance on OSWorld, CUA-World, and video game action prediction with 2.5-5.8$\times$ fewer visual tokens and up to 3.3$\times$ lower inference cost, that it outperforms other token reduction approaches, and that learning visual token selection jointly with the model is crucial.

\section{Visual Token Routing}
\label{sec:method}

\subsection{Problem Setup}
\label{sec:problem-setup}

Our goal is to reduce the number of visual tokens that a computer-use agent processes at each step, without degrading the actions that it takes. Next, we describe how such an agent observes and acts, and then formalize our method, Visual Token Routing (VTR).

\paragraph{Computer-use agents.} A computer-use agent performs a task specified by a natural-language instruction $p$ over multiple steps. At each step $t$, it receives an observation $o_t$, such as a screenshot of a desktop or a frame of a video game, and takes an action $a_t$, such as a mouse click or a key press. The agent chooses $a_t$ based on its interaction history $h_t = (p, o_1, a_1, \ldots, o_{t-1}, a_{t-1}, o_t)$, and we model it as a vision-language model $\pi_\theta$ that predicts $a_t \sim \pi_\theta(\cdot \mid h_t)$.

\paragraph{Visual tokens.} The vision encoder of $\pi_\theta$ maps each observation $o_\tau$ to $N$ visual tokens $v_{\tau,1}, \ldots, v_{\tau,N} \in \mathbb{R}^D$, one per image region, where $D$ is the hidden size of the language model. For instance, $N = 2{,}040$ for a 1080p screenshot in Qwen3.5-2B~\citep{qwen2026qwen35}. As each observation in $h_t$ contributes $N$ tokens, the language model processes $tN$ visual tokens at step $t$.

\paragraph{Visual token routing.} A router assigns each visual token a gate $g_{\tau,j} \in \{0, 1\}$, and the language model receives the routed history $\tilde{h}_t$, which contains $v_{\tau,j}$ only if $g_{\tau,j} = 1$, along with all text and action tokens. Dropped tokens are removed from the input sequence, so the language model does not process them. Our goal is to minimize the fraction of kept visual tokens,
\begin{equation}
    R_t = \frac{1}{tN} \sum_{\tau=1}^{t} \sum_{j=1}^{N} g_{\tau,j},
    \label{eq:keep-fraction}
\end{equation}
without degrading the prediction of $a_t$ compared with the full history $h_t$. We do not fix $R_t$ in advance, so the number of kept tokens can vary across observations and steps. However, learning the gates raises two challenges. First, the gates are discrete, so the training loss provides no gradient for choosing them. Second, the right trade-off between keeping fewer tokens and predicting accurately differs across inputs. Next, we describe VTR, which addresses both.

\subsection{Spatio-Temporal Router}
\label{sec:router}

Whether a visual token is redundant depends on the earlier observations in the history. We therefore base each routing decision on the same image region across the current and the $K$ previous observations. Formally, given visual token $v_{\tau,j}$, the router concatenates the visual tokens at the same position $j$,
\begin{equation}
    x_{\tau,j} = [v_{\tau-K,j}; \ldots; v_{\tau-1,j}; v_{\tau,j}],
    \label{eq:router-input}
\end{equation}
and predicts a keep probability $p_{\tau,j} = \sigma(f_\phi(x_{\tau,j}))$, where $f_\phi$ is a small MLP. Beyond recognizing regions that have not changed, the router learns which regions matter for the agent's actions through the objective in \S\ref{sec:objective}.

\paragraph{Gating.} The router samples a gate $g_{\tau,j} \sim \mathrm{Bernoulli}(p_{\tau,j})$, and a kept token enters the language model scaled by its keep probability,
\begin{equation}
    \tilde{v}_{\tau,j} = p_{\tau,j} \, v_{\tau,j} \quad \text{if } g_{\tau,j} = 1.
    \label{eq:gating}
\end{equation}
The scaling makes the language model's loss depend on $p_{\tau,j}$, which gives the router a gradient despite the discrete gate, as in mixture-of-experts models~\citep{shazeer2017moe,fedus2022switch}. Kept tokens retain their original positions, so the language model still knows where each kept region lies on the screen. At inference, the router runs inside vLLM~\citep{kwon2023vllm} and samples the gates as in training.

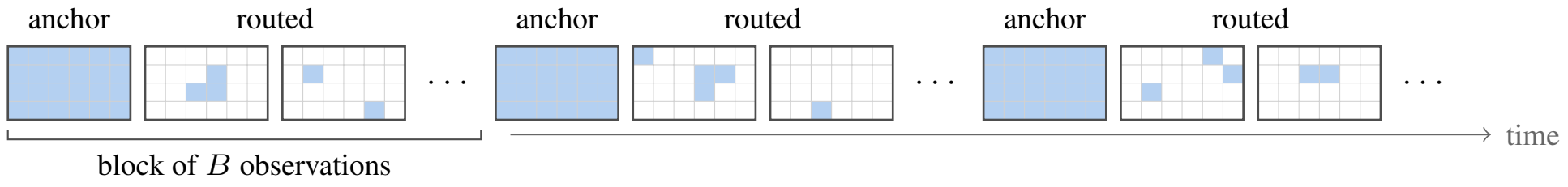
\begin{figure}[t]
\centering
\definecolor{vtr}{HTML}{2A78D6}
\newcommand{\obsframe}[2]{%
  \begin{scope}[shift={(#1,0)}]
    \foreach \i/\j in {#2} \fill[vtr!35] ({\i/6},{\j*0.15}) rectangle ++(1/6,0.15);
    \draw[black!18, line width=0.2pt, xstep=1/6, ystep=0.15] (0,0) grid (1,0.6);
    \draw[black!70, line width=0.5pt] (0,0) rectangle (1,0.6);
  \end{scope}}
\newcommand{\anchorframe}[1]{%
  \begin{scope}[shift={(#1,0)}]
    \fill[vtr!35] (0,0) rectangle (1,0.6);
    \draw[black!18, line width=0.2pt, xstep=1/6, ystep=0.15] (0,0) grid (1,0.6);
    \draw[black!70, line width=0.5pt] (0,0) rectangle (1,0.6);
  \end{scope}}
\begin{tikzpicture}[font=\scriptsize]
\anchorframe{0.00}
\obsframe{1.12}{2/1,3/1,3/2}
\obsframe{2.24}{1/2,4/0}
\node at (3.61,0.3) {$\cdots$};
\node[above] at (0.50,0.62) {anchor};
\node[above] at (2.18,0.62) {routed};
\anchorframe{3.98}
\obsframe{5.10}{3/1,3/2,4/2,0/3}
\obsframe{6.22}{2/0}
\node at (7.59,0.3) {$\cdots$};
\node[above] at (4.48,0.62) {anchor};
\node[above] at (6.16,0.62) {routed};
\anchorframe{7.96}
\obsframe{9.08}{1/1,4/3,5/2}
\obsframe{10.20}{2/2,3/2}
\node at (11.57,0.3) {$\cdots$};
\node[above] at (8.46,0.62) {anchor};
\node[above] at (10.14,0.62) {routed};
\draw[black!70, line width=0.4pt] (0,-0.08) -- (0,-0.16) -- (3.86,-0.16) -- (3.86,-0.08);
\node[below] at (1.93,-0.16) {block of $B$ observations};
\draw[->, black!60, line width=0.4pt] (4.1,-0.12) -- (12.1,-0.12) node[right] {time};
\end{tikzpicture}
\caption{\textbf{Where the router acts.} The first observation of each block is an anchor and is kept in full, while the router keeps only some regions, shown shaded, of the other observations in the block.}
\label{fig:anchors}
\end{figure}

\paragraph{Anchors.} We group observations into blocks of $B$ consecutive observations and call the first observation of each block an anchor (Figure~\ref{fig:anchors}). Anchors are kept in full, and the router acts only on the other observations in a block, with its history starting at the anchor. The anchor gives the model a full view of the screen to compare later observations against, and each new block refreshes this view, so regions that the router keeps dropping are never out of date for long.

\paragraph{Sliding router.} We also consider a sliding variant, in which only the first observation is kept in full, and the router's history is a rolling window of the $K$ previous observations without block boundaries. This variant avoids the cost of periodic anchors and uses the same trained router.

\subsection{Training}
\label{sec:objective}

\begin{figure}[t]
\centering
\includegraphics[width=\linewidth]{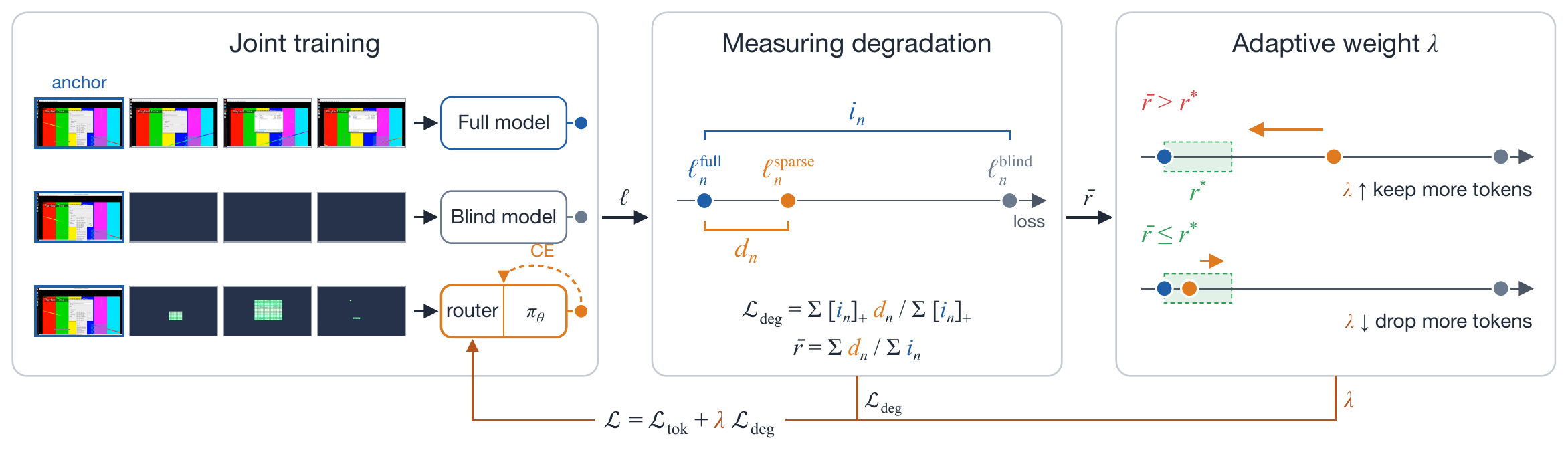}
\caption{\textbf{Training in VTR.} \textbf{Joint training:} We train $\pi_\theta$ and the router together with the cross-entropy (CE) loss on the kept tokens, and also train the router to keep fewer tokens with $\mathcal{L} = \mathcal{L}_{\mathrm{tok}} + \lambda\,\mathcal{L}_{\mathrm{deg}}$. The full and blind models see all tokens and only the anchors, and are used only during training. \textbf{Measuring degradation:} We compare the losses of the three models to measure how much the full history helps ($i_n$) and how much dropping tokens hurts ($d_n$). \textbf{Adaptive weight $\lambda$:} We increase $\lambda$ when the share of this help that $\pi_\theta$ loses, $\bar{r}$, exceeds the small tolerance $r^\star$, so that the router keeps more tokens, and decrease it otherwise.}
\label{fig:training}
\end{figure}

Training the router requires balancing two goals: keeping as few visual tokens as possible and preserving the accuracy of action prediction (Figure~\ref{fig:training}). A natural objective is a weighted sum of the two,
\begin{equation}
    \mathcal{L} = \mathcal{L}_{\mathrm{tok}} + \lambda \, \mathcal{L}_{\mathrm{deg}},
    \label{eq:objective}
\end{equation}
where $\mathcal{L}_{\mathrm{tok}}$ is the expected fraction of kept tokens, $\mathcal{L}_{\mathrm{deg}}$ measures how much dropping tokens hurts action prediction, and $\lambda$ is a weight. This raises two questions: how to measure $\mathcal{L}_{\mathrm{deg}}$, and how to set $\lambda$. For the first question, the loss of $\pi_\theta$ alone is not enough, since some actions are hard to predict even from the full history, and others can be predicted from the anchors alone. We therefore compare $\pi_\theta$ against two reference models: one that sees the full history and one that sees only the anchors. For the second question, the right $\lambda$ is unknown in advance, since a large $\lambda$ keeps too many tokens and a small $\lambda$ loses too much accuracy. We therefore set a target for how much accuracy the router may give up, measured as a fraction of how much the full history improves over the anchors alone, and adapt $\lambda$ during training to meet it.

\paragraph{Measuring degradation.} We call the two reference models the full model and the blind model. For each action token $n$, let $\ell^{\mathrm{full}}_n$, $\ell^{\mathrm{blind}}_n$ and $\ell^{\mathrm{sparse}}_n$ be the cross-entropy losses of the full model, the blind model and $\pi_\theta$ given only the kept tokens. We define
\begin{equation}
    i_n = \ell^{\mathrm{blind}}_n - \ell^{\mathrm{full}}_n, \qquad d_n = \ell^{\mathrm{sparse}}_n - \ell^{\mathrm{full}}_n,
    \label{eq:importance-degradation}
\end{equation}
so $i_n$ is how much the full history helps predict token $n$, and $d_n$ is how much dropping tokens hurts it. Subtracting $\ell^{\mathrm{full}}_n$ removes how hard token $n$ is to predict in the first place. We then average $d_n$ over action tokens, weighted by $i_n$,
\begin{equation}
    \mathcal{L}_{\mathrm{deg}} = \frac{\sum_n [i_n]_+ \, d_n}{\sum_n [i_n]_+},
    \label{eq:degradation}
\end{equation}
where $[x]_+ = \max(x, 0)$. Tokens that the anchors alone predict as well as the full history get zero weight, since dropping tokens should not matter for them, while tokens that depend most on the full history get the most weight. Finally, we compute $\mathcal{L}_{\mathrm{tok}}$ as the mean keep probability over the visual tokens outside the anchors, which, unlike the gates, is differentiable.

\paragraph{Adaptive weight.} Rather than fixing $\lambda$, we formulate training as a constrained optimization problem: minimize the expected fraction of kept tokens, subject to the model losing at most a fraction $r^\star$ of the improvement that the full history provides over the anchors,
\begin{equation}
    \min \; \mathcal{L}_{\mathrm{tok}} \quad \text{subject to} \quad \bar{r} \leq r^\star, \qquad \text{where} \quad \bar{r} = \frac{\sum_n d_n}{\sum_n i_n}.
    \label{eq:constrained}
\end{equation}
Eq.~\eqref{eq:objective} corresponds to the Lagrangian of this problem, with $\lambda$ as the Lagrange multiplier. We therefore update $\lambda$ by dual ascent~\citep{boyd2004convex}, alternating between a gradient step on the router and an update of $\lambda$,
\begin{equation}
    \lambda \leftarrow \max\bigl(0,\ \lambda + \alpha\,(\bar{r} - r^\star)\bigr),
    \label{eq:lambda-update}
\end{equation}
where $\bar{r}$ is computed over the training batch and $\alpha$ is the step size. In practice, $\lambda$ increases when $\bar{r}$ exceeds $r^\star$, placing more weight on $\mathcal{L}_{\mathrm{deg}}$ so that the router keeps more tokens, and decreases when $\bar{r}$ is below $r^\star$, so that the router drops more tokens. At a fixed point of this procedure, either $\bar{r} = r^\star$ or $\lambda = 0$, as required for a solution of Eq.~\eqref{eq:constrained}. Notably, $r^\star$ constrains the degradation rather than the number of kept tokens, so the router learns how many tokens each input needs.

\paragraph{Joint training.} Since $\pi_\theta$ was pre-trained on full images, we train it jointly with the router so that it learns to predict actions from the kept tokens alone. At each training step, we sample gates for each example and update $\pi_\theta$ and the router with the cross-entropy loss of $\pi_\theta$ given only the kept tokens. We then sample a second set of gates and update only the router with Eq.~\eqref{eq:objective}, keeping $\pi_\theta$ fixed. The full and blind models are initialized from the same model as $\pi_\theta$ and trained with the cross-entropy loss on the same examples, so that $i_n$ and $d_n$ reflect differences in the input rather than in training. At inference, we use only $\pi_\theta$ and the router, so the reference models add no cost.

\paragraph{Initialization.} In joint training, the router learns which tokens matter only through the loss of $\pi_\theta$, which gives a weak signal when the router starts from random weights. We therefore first train the router alone to predict simple labels based on \S\ref{sec:introduction}: a token is labeled as kept only if its region changed from the previous observation, so that it is not temporally redundant, and belongs to an object segmented by SAM~\citep{kirillov2023sam}, so that it is likely semantically relevant. Joint training then starts from this router and refines its decisions based on what the agent needs for its next action. Further details are in Appendix~\ref{app:initialization}.

\section{Experimental Setup}
\label{sec:experiments}

We evaluate VTR in two domains: computer-use tasks and video games.

\paragraph{Models.} We apply VTR to Qwen3.5-2B~\citep{qwen2026qwen35}. We train a single model on both desktop tasks and video games, and evaluate it in both domains. The router is a three-layer MLP with 23.3M parameters, about 1\% of the size of the language model, which reads each image region across the current and $K=10$ previous observations. We place an anchor every $B=10$ observations, and set the target $r^\star=0.1$ (\S\ref{sec:objective}). Further details are in Appendix~\ref{app:experimental-details}.

\paragraph{Desktop tasks.} We train on computer-use trajectories from CUA-World train, in which each screenshot is at a resolution of $1920\times1080$ giving 2,040 visual tokens. We evaluate the agent on OSWorld~\citep{xie2024osworldbenchmarkingmultimodalagents}, which contains tasks in common desktop applications such as web browsers and office software, and on CUA-World~\citep{aggarwal2026gymanythingturnsoftwareagent}, which contains tasks across 200 professional software applications.

\paragraph{Video games.} We train on gameplay recordings at 30 frames per second from the \texttt{markov-ai/gaming-500-hours} dataset, with each frame resized to $640\times360$, which gives 220 visual tokens. We evaluate how accurately the agent predicts the next action, such as a click or a key press, on held-out recordings from 36 games.

\paragraph{Baselines.} We compare VTR with the following ways to reduce visual tokens, all using the same model and training data:
\begin{enumerate}[leftmargin=*,nosep]
    \item \textbf{Lower resolution:} the dense model, which processes every visual token, with each image resized to 50\%, 25\%, or 12.5\% of its area.
    \item \textbf{Pixel-diff:} keeps only the regions whose pixels changed from the previous observation, following methods that drop tokens repeated across video frames~\citep{choudhury2024rlt,bagrov2025evs} and screenshots~\citep{abaskohi2026revision}.
    \item \textbf{H.264:} keeps only the regions that the H.264 video codec~\citep{wiegand2003h264} encodes as moved or changed, following methods that select visual tokens from the motion vectors and residuals of video codecs~\citep{tang2026ovencoder,sarkar2026cope}.
    \item \textbf{Frozen router:} the initial router of VTR, kept fixed while training the model, which shows the effect of training the router jointly with the model.
\end{enumerate}

As in VTR, both rules keep the anchors whole.
Each baseline is trained with same experimental setup and common hyperparameters, as VTR.
We also evaluate VTR with the sliding router (\S\ref{sec:router}), and with images resized to half or a quarter of their area.

\section{Results}
\label{sec:results}

\subsection{Main Results}
\label{sec:main-results}

\begin{figure}[t]
\centering
\definecolor{dense}{HTML}{8C949F}
\definecolor{vtr}{HTML}{2A78D6}
\definecolor{hxx}{HTML}{EB6834}
\definecolor{pdiff}{HTML}{1BAF7A}
\definecolor{frozen}{HTML}{9B59B6}
\begin{tikzpicture}
\begin{groupplot}[
    group style={group size=3 by 1, horizontal sep=1.0cm},
    width=0.345\linewidth, height=4.4cm,
    xmode=log, log basis x=10,
    axis x line*=bottom, axis y line*=left,
    grid=major, grid style={line width=0.3pt, draw=black!12},
    tick style={draw=none},
    tick label style={font=\footnotesize}, label style={font=\footnotesize},
    title style={font=\small, yshift=-4pt},
    xlabel={Visual tokens per image},
    log ticks with fixed point,
    every axis plot/.append style={line width=1pt, mark size=2.2pt},
]
\nextgroupplot[title={OSWorld}, ylabel={Score}, xmin=130, xmax=2600, xtick={200,500,1000,2000}, ymin=0, ymax=22,
    legend to name=mainresultslegend, legend columns=6,
    legend style={draw=none, font=\footnotesize, /tikz/every even column/.append style={column sep=5pt}}]
\addplot[dense, mark=*] coordinates {(2040,18.56) (1008,15.19) (510,13.39) (252,6.17)};
\addplot[vtr, mark=square*] coordinates {(807.9,18.10) (391,13.73) (171.8,5.76)};
\addplot[vtr, mark=square, only marks, mark options={line width=1pt}] coordinates {(712.6,18.10)};
\addplot[hxx, mark=triangle*, only marks, mark size=2.8pt] coordinates {(1386.5,9.44)};
\addplot[pdiff, mark=diamond*, only marks, mark size=2.8pt] coordinates {(485.9,5.25)};
\addplot[frozen, mark=pentagon*, only marks, mark size=2.6pt] coordinates {(421.5,4.91)};
\legend{Dense (fixed resolution), VTR (ours), VTR sliding (ours), H.264, Pixel-diff, Frozen router}
\draw[dense, dashed, line width=0.6pt] (axis cs:130,18.56) -- (axis cs:2600,18.56);
\nextgroupplot[title={CUA-World}, ylabel={Average score}, xmin=110, xmax=2500, xtick={200,500,1000,2000}, ymin=0, ymax=20]
\addplot[dense, mark=*] coordinates {(1946.2,16.61) (966.5,15.33) (510,11.75) (252,8.98)};
\addplot[vtr, mark=square*] coordinates {(674.3,16.34) (328.8,12.92) (148.1,9.25)};
\addplot[vtr, mark=square, only marks, mark options={line width=1pt}] coordinates {(595.2,15.77)};
\addplot[hxx, mark=triangle*, only marks, mark size=2.8pt] coordinates {(1367,12.43)};
\addplot[pdiff, mark=diamond*, only marks, mark size=2.8pt] coordinates {(406.8,9.33)};
\draw[dense, dashed, line width=0.6pt] (axis cs:110,16.61) -- (axis cs:2500,16.61);
\nextgroupplot[title={Video games (click)}, ylabel={Click accuracy (\%)}, xmin=7, xmax=280, xtick={10,30,100,200}, ymin=0, ymax=6]
\addplot[dense, mark=*] coordinates {(220,4.74) (112,3.82) (55,3.63) (28,3.44)};
\addplot[vtr, mark=square*] coordinates {(38.2,4.90) (16.4,4.54) (9.5,4.31)};
\addplot[vtr, mark=square, only marks, mark options={line width=1pt}] coordinates {(23.2,4.34)};
\addplot[hxx, mark=triangle*, only marks, mark size=2.8pt] coordinates {(150.3,2.54)};
\addplot[pdiff, mark=diamond*, only marks, mark size=2.8pt] coordinates {(67.3,4.59)};
\addplot[frozen, mark=pentagon*, only marks, mark size=2.6pt] coordinates {(51.6,4.14)};
\draw[dense, dashed, line width=0.6pt] (axis cs:7,4.74) -- (axis cs:280,4.74);
\end{groupplot}
\node[anchor=south] at ($(group c1r1.north west)!0.5!(group c3r1.north east)$) [yshift=12pt] {\pgfplotslegendfromname{mainresultslegend}};
\end{tikzpicture}
\caption{\textbf{VTR matches full-resolution performance with far fewer visual tokens.} Score on OSWorld (left), average score on CUA-World (middle), and click accuracy in video games (right), against visual tokens per image. The dashed line marks the dense model at full resolution.}
\label{fig:main-results}
\end{figure}
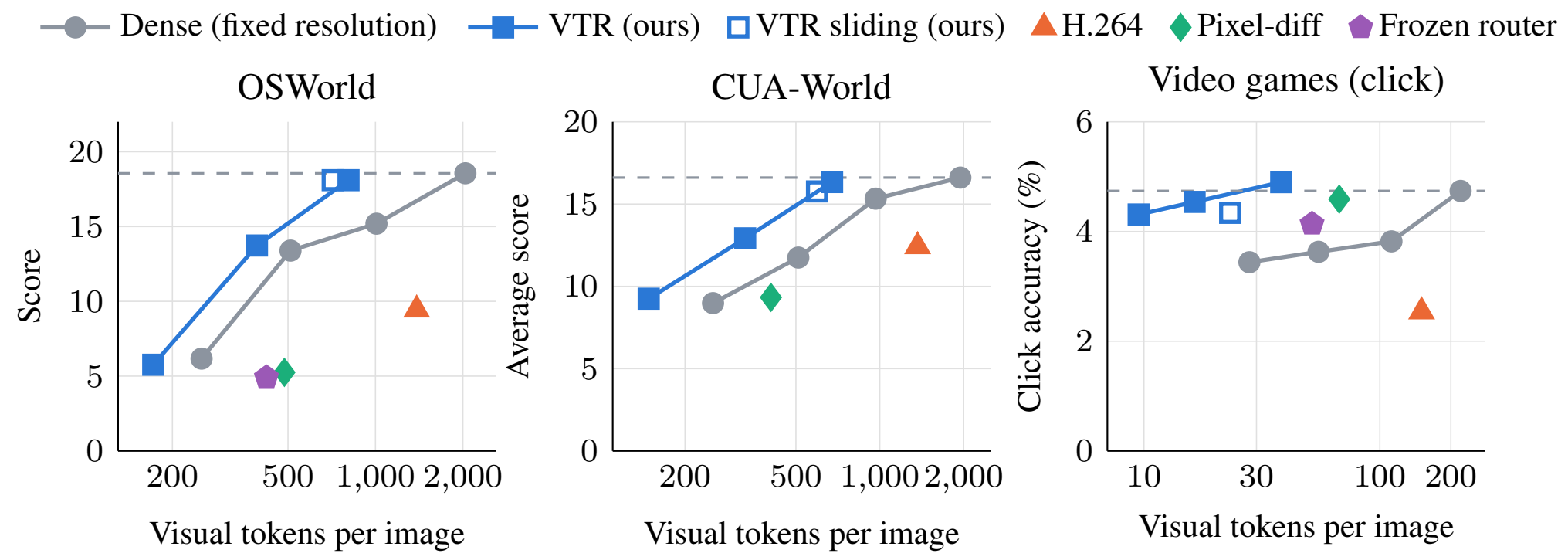

\paragraph{VTR matches full-resolution performance with far fewer visual tokens.} Figure~\ref{fig:main-results} compares VTR with the dense model, which processes every visual token, across numbers of tokens per image. On OSWorld and CUA-World, VTR matches the dense model at full resolution while using 2.5$\times$ and 2.9$\times$ fewer visual tokens, and in video games, it uses 5.8$\times$ fewer. Interestingly, the same router keeps about 40\% of the tokens on OSWorld but only 17\% in video games. Since we never specify how many tokens to keep (\S\ref{sec:objective}), the router learns this difference on its own, likely because consecutive game frames are only 1/30 of a second apart, while desktop screenshots are taken after each action. These results show that most visual tokens in an agent's history can be removed without degrading its actions, and that the router adapts how many it removes to the input.

\paragraph{VTR outperforms other token reduction approaches.} Figure~\ref{fig:main-results} also compares VTR with two other ways to reduce visual tokens: lowering the image resolution, and hand-designed rules that keep the regions that changed between observations, namely pixel-diff and H.264. At every number of tokens where both are measured, VTR achieves higher performance than the dense model at a lower resolution. For instance, on OSWorld, VTR scores 2.9 points higher than the dense model at half the image area, while using fewer tokens. Lowering the resolution removes detail from every region of the screen, including the small regions that actions depend on, like a cursor or a spreadsheet cell, whereas VTR keeps the regions it selects at full resolution. VTR also outperforms both hand-designed rules on every benchmark. On OSWorld, for instance, VTR scores 18.1, compared with 9.4 for H.264 and 5.3 for pixel-diff. Unlike these rules, which decide what to keep from the pixels alone, VTR learns what the agent needs for its next action in each domain. Moreover, VTR is complementary to lowering the resolution: combined with a quarter of the image area, it reduces visual tokens by 23$\times$ in video games, with only a small drop in click accuracy. These results show that learning which regions to keep preserves more of what the agent needs than lowering the resolution or applying a fixed rule.

\paragraph{Learning which tokens to keep jointly with the model is crucial.} To isolate the effect of joint training, we compare VTR with a frozen router, which starts from the same initial router (\S\ref{sec:objective}) but keeps it fixed while training the language model. On OSWorld, VTR scores 18.1, compared with only 4.9 for the frozen router, which performs similarly to pixel-diff. This is expected, since the initial router is trained on a similar rule, keeping regions that changed and belong to an object, and training the language model alone cannot recover the regions that this router drops. In video games, VTR also achieves higher click accuracy than the frozen router, although the gap is smaller. These results show that the advantage of VTR comes from training the router jointly with the model, rather than from the initial rule.

\subsection{Analysis}
\label{sec:analysis}

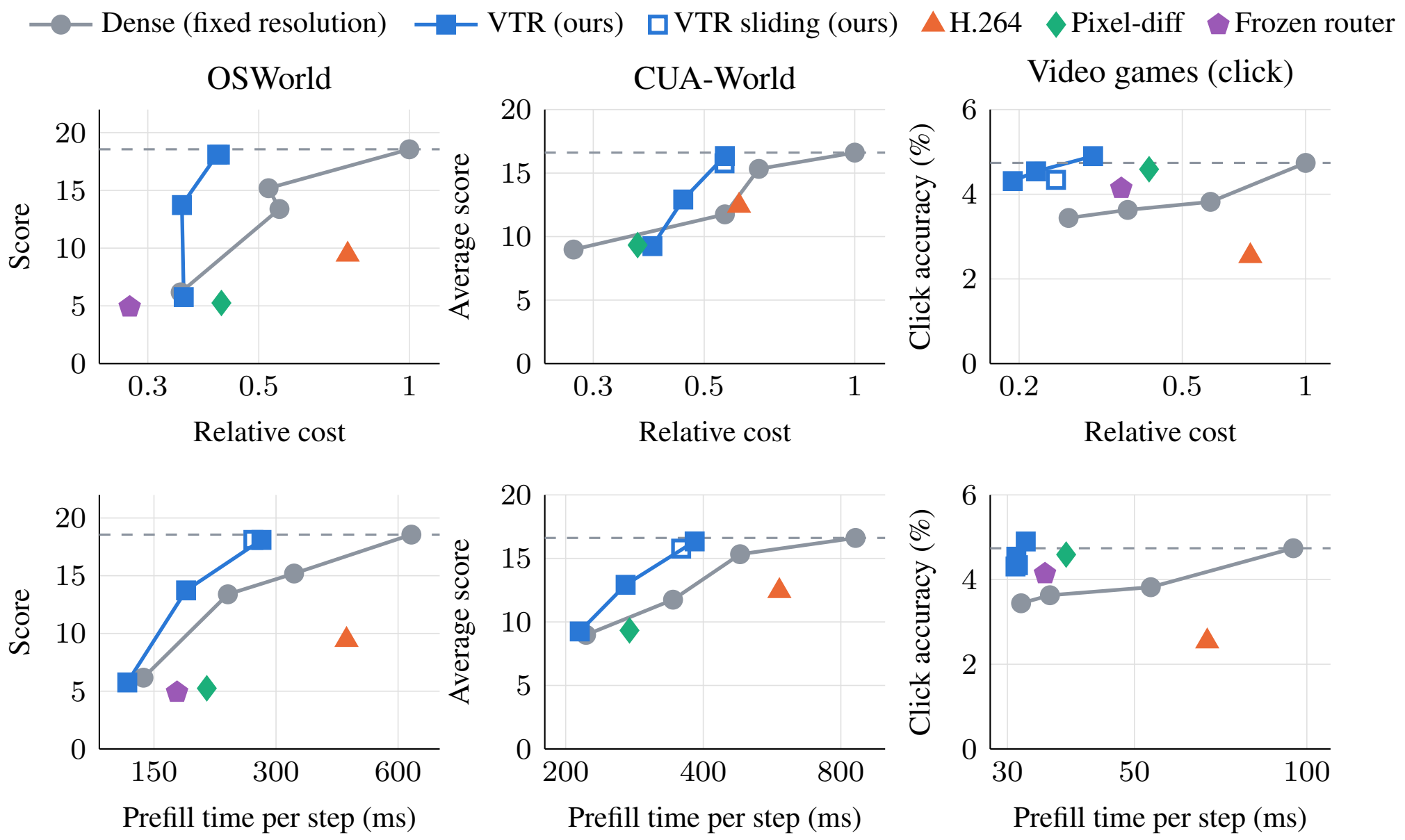
\begin{figure}[t]
\centering
\definecolor{dense}{HTML}{8C949F}
\definecolor{vtr}{HTML}{2A78D6}
\definecolor{hxx}{HTML}{EB6834}
\definecolor{pdiff}{HTML}{1BAF7A}
\definecolor{frozen}{HTML}{9B59B6}
\begin{tikzpicture}
\begin{groupplot}[
    group style={group size=3 by 2, horizontal sep=1.0cm, vertical sep=1.25cm},
    width=0.345\linewidth, height=4.0cm,
    xmode=log, log basis x=10,
    axis x line*=bottom, axis y line*=left,
    grid=major, grid style={line width=0.3pt, draw=black!12},
    tick style={draw=none},
    tick label style={font=\footnotesize}, label style={font=\footnotesize},
    title style={font=\small, yshift=-4pt},
    log ticks with fixed point,
    every axis plot/.append style={line width=1pt, mark size=2.2pt},
]
\nextgroupplot[title={OSWorld}, ylabel={Score}, xlabel={Relative cost}, xmin=0.24, xmax=1.15, xtick={0.3,0.5,1}, ymin=0, ymax=22,
    legend to name=costspeedlegend, legend columns=6,
    legend style={draw=none, font=\footnotesize, /tikz/every even column/.append style={column sep=5pt}}]
\addplot[dense, mark=*] coordinates {(1.0,18.56) (0.523,15.19) (0.55,13.39) (0.349,6.17)};
\addplot[vtr, mark=square*] coordinates {(0.413,18.10) (0.351,13.73) (0.354,5.76)};
\addplot[vtr, mark=square, only marks, mark options={line width=1pt}] coordinates {(0.418,18.10)};
\addplot[hxx, mark=triangle*, only marks, mark size=2.8pt] coordinates {(0.753,9.44)};
\addplot[pdiff, mark=diamond*, only marks, mark size=2.8pt] coordinates {(0.421,5.25)};
\addplot[frozen, mark=pentagon*, only marks, mark size=2.6pt] coordinates {(0.276,4.91)};
\legend{Dense (fixed resolution), VTR (ours), VTR sliding (ours), H.264, Pixel-diff, Frozen router}
\draw[dense, dashed, line width=0.6pt] (axis cs:0.24,18.56) -- (axis cs:1.15,18.56);
\nextgroupplot[title={CUA-World}, ylabel={Average score}, xlabel={Relative cost}, xmin=0.24, xmax=1.15, xtick={0.3,0.5,1}, ymin=0, ymax=20]
\addplot[dense, mark=*] coordinates {(1.0,16.61) (0.643,15.33) (0.55,11.75) (0.274,8.98)};
\addplot[vtr, mark=square*] coordinates {(0.55,16.34) (0.454,12.92) (0.394,9.25)};
\addplot[vtr, mark=square, only marks, mark options={line width=1pt}] coordinates {(0.549,15.77)};
\addplot[hxx, mark=triangle*, only marks, mark size=2.8pt] coordinates {(0.587,12.43)};
\addplot[pdiff, mark=diamond*, only marks, mark size=2.8pt] coordinates {(0.368,9.33)};
\draw[dense, dashed, line width=0.6pt] (axis cs:0.24,16.61) -- (axis cs:1.15,16.61);
\nextgroupplot[title={Video games (click)}, ylabel={Click accuracy (\%)}, xlabel={Relative cost}, xmin=0.17, xmax=1.15, xtick={0.2,0.5,1}, ymin=0, ymax=6]
\addplot[dense, mark=*] coordinates {(1.0,4.74) (0.586,3.82) (0.368,3.63) (0.264,3.44)};
\addplot[vtr, mark=square*] coordinates {(0.303,4.90) (0.22,4.54) (0.193,4.31)};
\addplot[vtr, mark=square, only marks, mark options={line width=1pt}] coordinates {(0.246,4.34)};
\addplot[hxx, mark=triangle*, only marks, mark size=2.8pt] coordinates {(0.733,2.54)};
\addplot[pdiff, mark=diamond*, only marks, mark size=2.8pt] coordinates {(0.415,4.59)};
\addplot[frozen, mark=pentagon*, only marks, mark size=2.6pt] coordinates {(0.355,4.14)};
\draw[dense, dashed, line width=0.6pt] (axis cs:0.17,4.74) -- (axis cs:1.15,4.74);
\nextgroupplot[ylabel={Score}, xlabel={Prefill time per step (ms)}, xmin=110, xmax=760, xtick={150,300,600}, ymin=0, ymax=22]
\addplot[dense, mark=*] coordinates {(647.1,18.56) (332.4,15.19) (228.2,13.39) (141.3,6.17)};
\addplot[vtr, mark=square*] coordinates {(276.2,18.10) (180.0,13.73) (128.9,5.76)};
\addplot[vtr, mark=square, only marks, mark options={line width=1pt}] coordinates {(263.8,18.10)};
\addplot[hxx, mark=triangle*, only marks, mark size=2.8pt] coordinates {(447.7,9.44)};
\addplot[pdiff, mark=diamond*, only marks, mark size=2.8pt] coordinates {(202.5,5.25)};
\addplot[frozen, mark=pentagon*, only marks, mark size=2.6pt] coordinates {(171.0,4.91)};
\draw[dense, dashed, line width=0.6pt] (axis cs:110,18.56) -- (axis cs:760,18.56);
\nextgroupplot[ylabel={Average score}, xlabel={Prefill time per step (ms)}, xmin=180, xmax=1000, xtick={200,400,800}, ymin=0, ymax=20]
\addplot[dense, mark=*] coordinates {(861.6,16.61) (481.5,15.33) (343.5,11.75) (221.6,8.98)};
\addplot[vtr, mark=square*] coordinates {(383.1,16.34) (270.5,12.92) (214.7,9.25)};
\addplot[vtr, mark=square, only marks, mark options={line width=1pt}] coordinates {(357.8,15.77)};
\addplot[hxx, mark=triangle*, only marks, mark size=2.8pt] coordinates {(587.0,12.43)};
\addplot[pdiff, mark=diamond*, only marks, mark size=2.8pt] coordinates {(275.8,9.33)};
\draw[dense, dashed, line width=0.6pt] (axis cs:180,16.61) -- (axis cs:1000,16.61);
\nextgroupplot[ylabel={Click accuracy (\%)}, xlabel={Prefill time per step (ms)}, xmin=28, xmax=110, xtick={30,50,100}, ymin=0, ymax=6]
\addplot[dense, mark=*] coordinates {(94.7,4.74) (53.4,3.82) (35.6,3.63) (31.7,3.44)};
\addplot[vtr, mark=square*] coordinates {(32.3,4.90) (31.1,4.54) (31.0,4.31)};
\addplot[vtr, mark=square, only marks, mark options={line width=1pt}] coordinates {(31.3,4.34)};
\addplot[hxx, mark=triangle*, only marks, mark size=2.8pt] coordinates {(67.0,2.54)};
\addplot[pdiff, mark=diamond*, only marks, mark size=2.8pt] coordinates {(38.0,4.59)};
\addplot[frozen, mark=pentagon*, only marks, mark size=2.6pt] coordinates {(34.9,4.14)};
\draw[dense, dashed, line width=0.6pt] (axis cs:28,4.74) -- (axis cs:110,4.74);
\end{groupplot}
\node[anchor=south] at ($(group c1r1.north west)!0.5!(group c3r1.north east)$) [yshift=12pt] {\pgfplotslegendfromname{costspeedlegend}};
\end{tikzpicture}
\caption{\textbf{VTR makes inference cheaper and faster.} Score on OSWorld (left), average score on CUA-World (middle), and click accuracy in video games (right), against inference cost relative to the dense model at full resolution (top) and prefill time per step (bottom). The dashed line marks the dense model at full resolution.}
\label{fig:cost-speed}
\end{figure}

\paragraph{VTR makes inference cheaper and faster.} Since VTR removes most visual tokens, each step of the agent also costs less and takes less time. Figure~\ref{fig:cost-speed} shows the performance of each method against its inference cost and its prefill time, the time the model takes to process its input before generating an action. VTR reduces the cost by 2.4$\times$ on OSWorld, 1.8$\times$ on CUA-World, and 3.3$\times$ in video games, and speeds up prefill by 2.3$\times$, 2.2$\times$, and 2.9$\times$. These gains are smaller than the reduction in visual tokens, since VTR removes only visual tokens, while the instruction, the previous actions, and the generated action stay the same. Moreover, VTR is both cheaper and faster than the dense model at half the image area in all three domains, while achieving higher performance. These results show that the token savings of VTR translate into lower cost and faster inference, while matching full-resolution performance.

\begin{wrapfigure}{r}{0.5\linewidth}
\vspace{-12pt}
\centering
\definecolor{vtr}{HTML}{2A78D6}
\definecolor{hxx}{HTML}{EB6834}
\definecolor{pdiff}{HTML}{1BAF7A}
\begin{tikzpicture}
\begin{axis}[
    width=\linewidth, height=4.4cm,
    ybar, bar width=4.5pt,
    enlarge x limits=0.14,
    symbolic x coords={desktop,low,medium,high},
    xtick=data,
    xticklabels={Desktop, Low\\motion\\games, Medium\\motion\\games, High\\motion\\games},
    xticklabel style={align=center},
    ymode=log, log origin=infty, ymin=1, ymax=100,
    ytick={1,3,10,30,100}, yticklabels={1,3,10,30,100},
    ylabel={Visual tokens kept (\%)},
    axis x line*=bottom, axis y line*=left,
    ymajorgrids, grid style={line width=0.3pt, draw=black!12},
    tick style={draw=none},
    tick label style={font=\footnotesize}, label style={font=\footnotesize},
    legend style={draw=none, font=\footnotesize, at={(0.5,1.03)}, anchor=south, legend columns=3,
        /tikz/every even column/.append style={column sep=4pt}},
    legend image code/.code={\draw[#1, draw=none] (0cm,-0.1cm) rectangle (0.25cm,0.1cm);},
]
\addplot[fill=vtr, draw=none] coordinates {(desktop,26.3) (low,2.7) (medium,5.0) (high,5.9)};
\addplot[fill=hxx, draw=none] coordinates {(desktop,64.3) (low,32) (medium,64) (high,70)};
\addplot[fill=pdiff, draw=none] coordinates {(desktop,9.6) (low,5.9) (medium,16) (high,26)};
\legend{VTR (ours), H.264, Pixel-diff}
\end{axis}
\end{tikzpicture}
\caption{\textbf{VTR adapts how many visual tokens it keeps to each domain and game.} Percentage of visual tokens kept by each method on desktop tasks and on video games grouped by how much of the screen changes between consecutive frames.}
\label{fig:keep-rate}
\vspace{-10pt}
\end{wrapfigure}
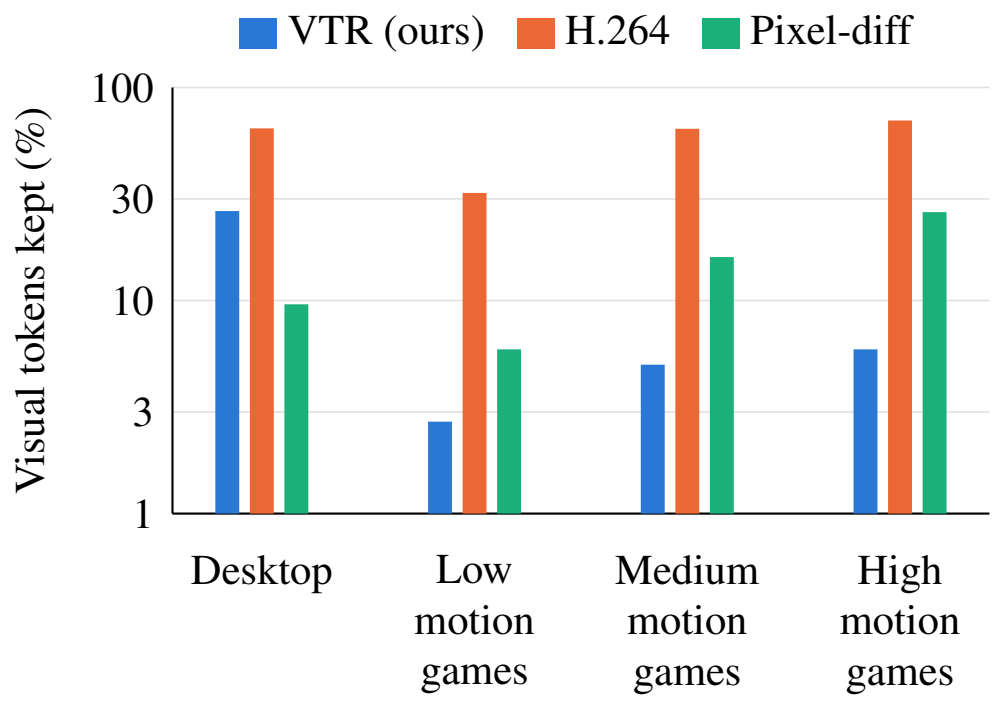

\paragraph{VTR adapts how many visual tokens it keeps to each domain and game.} Since we never fix the number of kept tokens (\S\ref{sec:objective}), the router decides how many tokens each input needs. To see how this number varies, we group the video games by how much of the screen changes between consecutive frames, from games like Candy Crush, where little of the screen moves, to games like Counter-Strike 2, where most of it does. Figure~\ref{fig:keep-rate} shows the percentage of visual tokens each method keeps. VTR keeps 2.7\% of the tokens in games with low motion, 5.9\% in games with high motion, and 26.3\% on desktop tasks. Interestingly, as more of a game screen changes, VTR keeps only a few more tokens, while pixel-diff and H.264 keep many more. This is likely because many regions that change, such as the background moving as the camera turns, do not affect the agent's next action. On desktop tasks, in contrast, VTR keeps more tokens than pixel-diff, likely because the agent often needs regions that did not change, such as text it has to read. These results show that VTR decides how many visual tokens to keep based on what each input needs, rather than on how much of the screen changes.

\begin{figure}[t]
\centering
\setlength{\tabcolsep}{1.5pt}
\renewcommand{\arraystretch}{0.6}
\begin{tabular}{@{}c cccc@{}}
& {\small Dota 2} & {\small Candy Crush} & {\small Minecraft} & {\small Gpredict} \\[3pt]
\rotatebox{90}{\hspace{6pt}\scriptsize Previous} &
\includegraphics[width=0.238\linewidth]{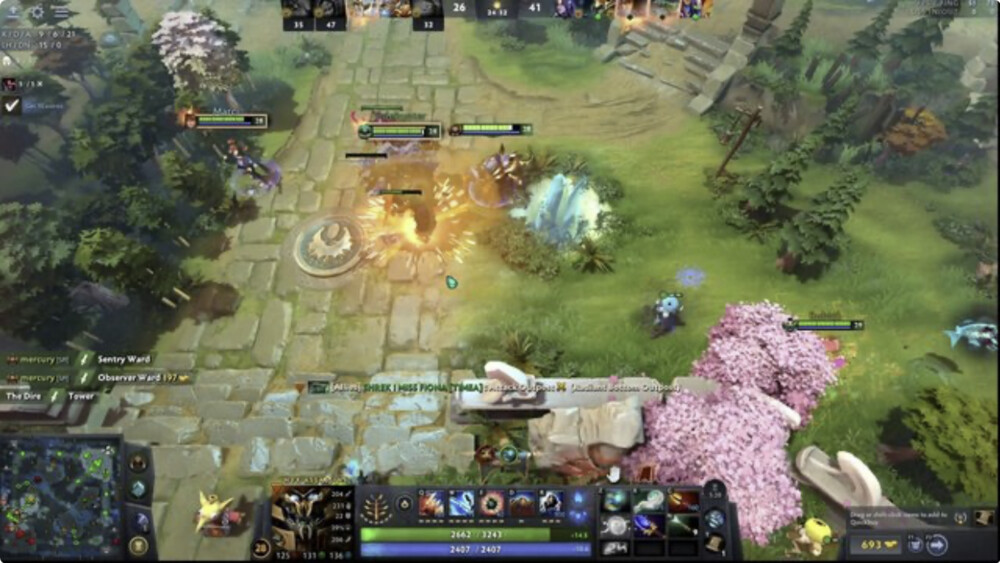} &
\includegraphics[width=0.238\linewidth]{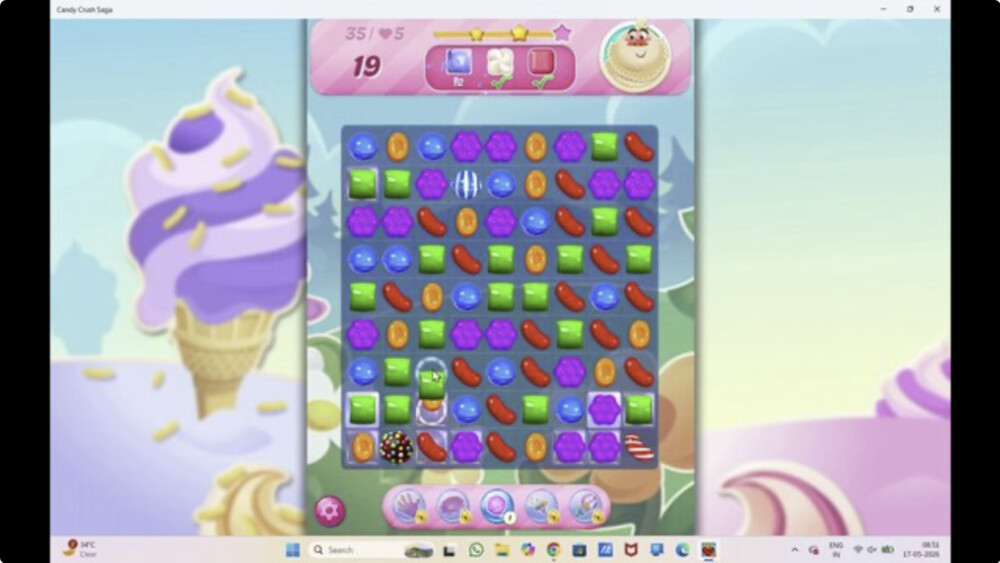} &
\includegraphics[width=0.238\linewidth]{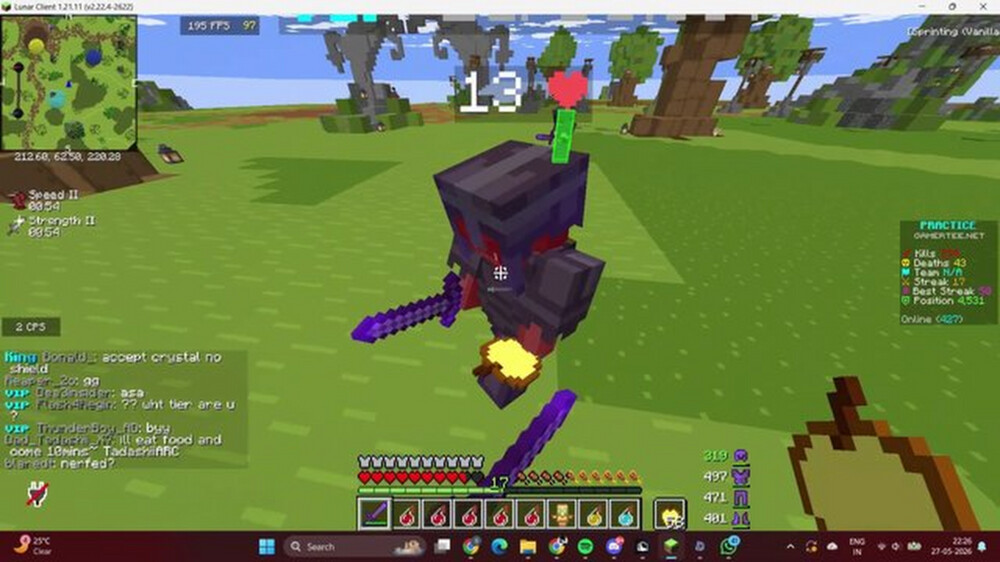} &
\includegraphics[width=0.238\linewidth]{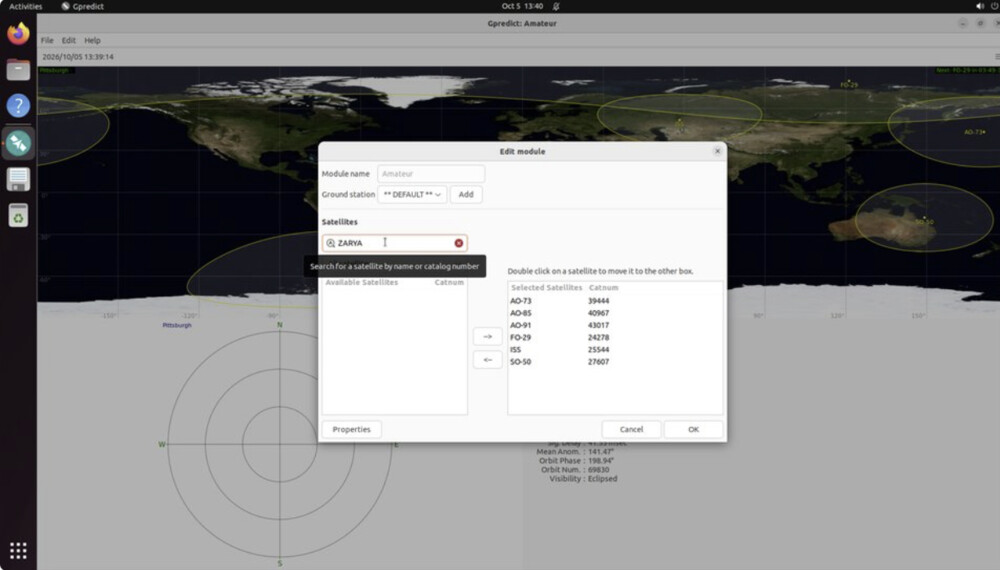} \\
\rotatebox{90}{\hspace{2pt}\scriptsize Kept by VTR} &
\includegraphics[width=0.238\linewidth]{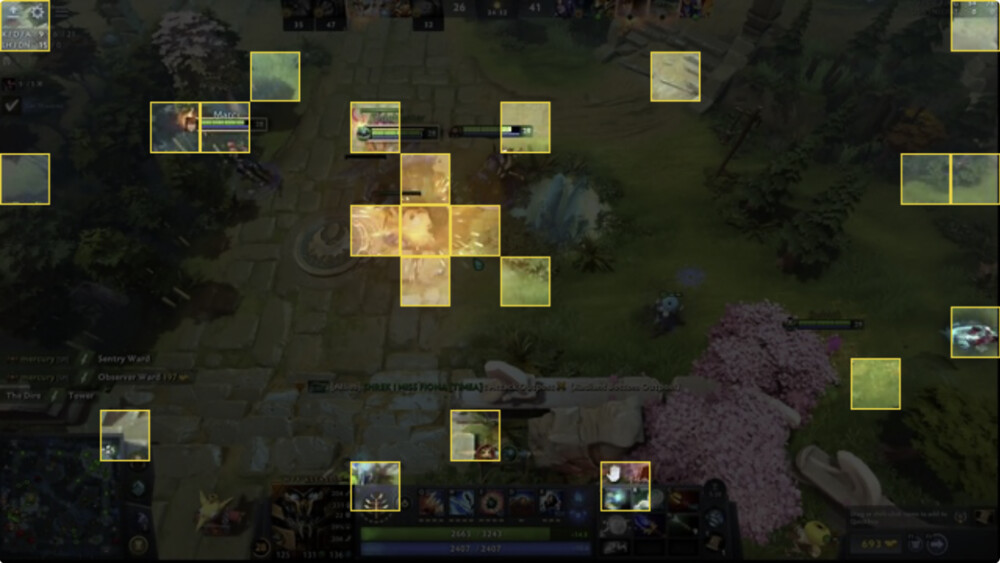} &
\includegraphics[width=0.238\linewidth]{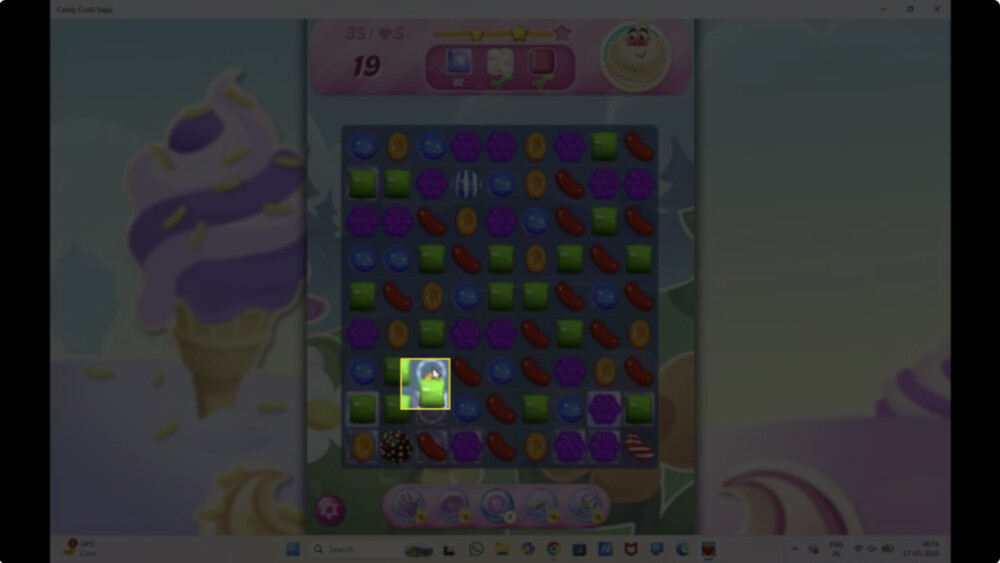} &
\includegraphics[width=0.238\linewidth]{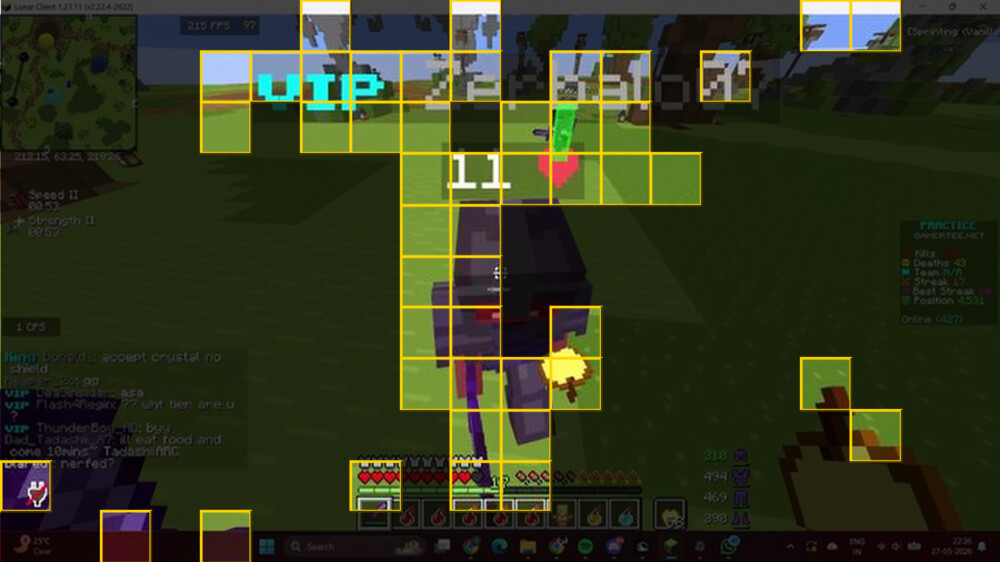} &
\includegraphics[width=0.238\linewidth]{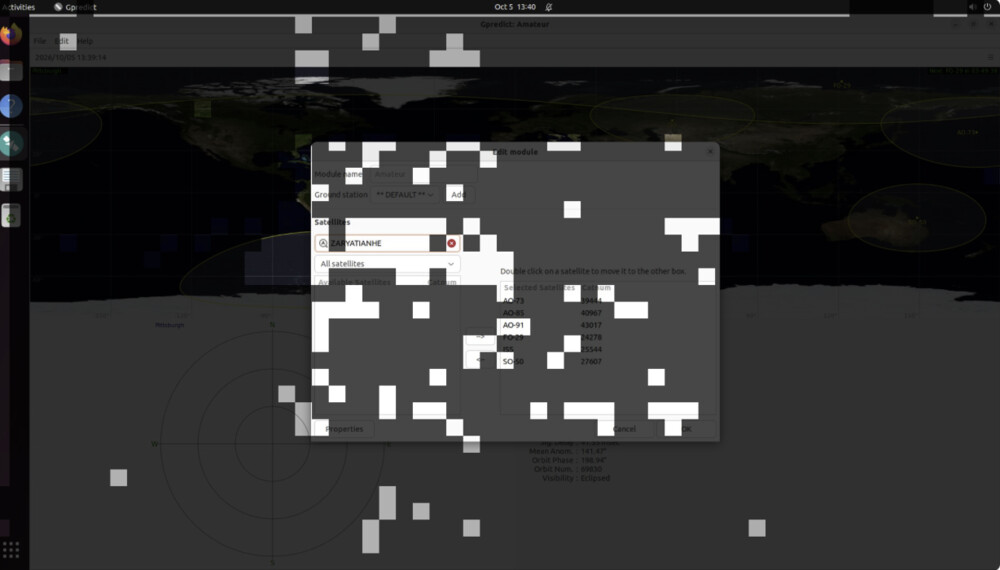} \\
\end{tabular}
\caption{\textbf{VTR keeps only the regions that the agent needs.} The previous observation (top) and the current observation with the regions that VTR drops dimmed (bottom), in three video games and a desktop application.}
\label{fig:qualitative}
\end{figure}

\paragraph{VTR keeps the regions the agent needs for its next action.} To see which regions the router keeps, Figure~\ref{fig:qualitative} shows the previous observation and the regions that VTR keeps in the current one. In Dota 2, a fight takes place on an otherwise static map, and VTR keeps 23 of the 220 regions, mostly the fighting units and their health bars. In Candy Crush, the player moves a single candy, and VTR keeps only the region of that candy. In Minecraft, VTR keeps the opponent and the health counter that dropped from 13 to 11, and drops the field around them. Interestingly, in Gpredict, VTR keeps the region where a tooltip disappeared, so it also treats the removal of an element as information, while it drops the satellite map behind the dialog. VTR also keeps a few isolated regions in static areas, likely because each gate is sampled from its keep probability. Overall, rather than just detecting motion, the router appears to keep the changes that the model cannot infer from earlier observations, such as a lower health counter or a closed tooltip, and to drop those it can, such as the shifted background when the camera moves in GTA V (Appendix~\ref{app:qualitative}). Further examples are in Appendix~\ref{app:qualitative}.

\section{Related Work}
\label{sec:related}

\paragraph{Visual token reduction.} Many methods reduce the visual tokens of vision-language models without training, by keeping the tokens that receive the most attention~\citep{chen2024fastv,yang2025visionzip,zhang2025sparsevlm}, merging similar tokens~\citep{bolya2023tome,shang2025prumerge}, or dropping tokens that repeat across video frames, detected from pixels or from video codecs~\citep{shen2025longvu,bagrov2025evs,choudhury2024rlt,tang2026ovencoder,sarkar2026cope}. Other methods train a module that predicts which tokens to keep~\citep{rao2021dynamicvit,huang2025dynamicllava,ye2025atpllava,wang2026score}. However, all of these methods set the number of kept tokens in advance, as a fixed ratio, a fixed threshold, or a target token count. In contrast, VTR trains its router jointly with the language model and keeps as many tokens as the model needs to preserve its accuracy.

\paragraph{Efficient computer-use agents.} Computer-use agents take far longer than humans to complete tasks~\citep{abhyankar2026osworldhumanbenchmarkingefficiencycomputeruse,aggarwal2026cuaspeedrun}, and latency limits them in real-time domains such as video games~\citep{zhang2025videogamebench,aggarwal2026gymanythingrt}. To reduce the cost of screenshots, agents commonly keep only the last few screenshots in context~\citep{qin2025uitars,wang2025opencua,anthropic2026computeruse}, and methods designed for GUI agents skip, prune, or compress visual tokens~\citep{lin2025showui,xu2026guipruner,huang2026guikv,chen2025simpagent,wang2026stpgui,abaskohi2026revision}. However, these methods either discard older screenshots entirely, or decide what to drop with fixed rules or a selector trained separately from the model that predicts actions. In contrast, VTR learns which tokens to keep together with the model, based on what the agent needs for its next action.

\paragraph{Learned routing and constrained optimization.} Mixture-of-experts and Mixture-of-Depths models learn to route tokens through parts of the network~\citep{shazeer2017moe,fedus2022switch,zhou2022expertchoice,raposo2024mod}, and adaptive computation methods let the amount of computation vary across inputs~\citep{graves2016act,wang2025remoe,schuster2022calm}. Separately, constrained optimization adapts a Lagrange multiplier during training to meet a target, for example in VAEs, reinforcement learning, and pruning~\citep{rezende2018geco,tessler2019rcpo,wang2020structured,gallegoposada2022controlled}. However, these methods fix the amount of computation, set a target for it, or calibrate it only after training, rather than constraining the model to keep its accuracy during training. In contrast, VTR minimizes the number of kept tokens subject to the model not losing accuracy, and enforces this constraint during training by dual ascent. Appendix~\ref{app:related} provides an extended discussion of related work.

\section{Conclusion}
\label{sec:conclusion}

In this work, we introduced Visual Token Routing (VTR), a joint training framework that enables computer-use agents to process only the visual tokens they need. By training a lightweight spatio-temporal router together with the vision-language model to keep as few visual tokens as possible without losing accuracy, VTR addresses a key computational bottleneck of computer-use agents, which arises from representing each screenshot with thousands of visual tokens. VTR matches the performance of the full-resolution model on OSWorld, CUA-World, and video game action prediction with 2.5-5.8$\times$ fewer visual tokens, reducing inference cost by up to 3.3$\times$ and speeding up prefill by up to 2.9$\times$, and outperforms lowering the image resolution and other baselines for dropping tokens. We further show that learning which tokens to keep jointly with the model is crucial. Our analysis also shows that the router learns on its own how many tokens to keep for each input and domain, keeping only regions that matter for the agent's next action. By learning what an agent needs to see, VTR makes it possible to use computer-use agents in long-horizon tasks and real-time domains at a fraction of the computational cost.

\section*{Acknowledgements}

We thank the CMU Foundation and Language Model (FLAME) Center for providing computational resources. We thank Jing Yu Koh for their feedback and helpful discussions. Pranjal Aggarwal was supported by SoftBank Group-Arm Fellowship during the project. This work is partially supported by the National Science Foundation under Grants DMS-2502281, DMS-2434614, and by DARPA ExpMath Grant HR0011262E028.

\bibliographystyle{plainnat}
\bibliography{references}

\clearpage
\appendix
\section{Experimental Details}
\label{app:experimental-details}

\subsection{Training Data}
The training data contains 4,265 computer-use examples from the CUA-World training split~\citep{aggarwal2026gymanythingturnsoftwareagent} and 29,401 video game examples from the \texttt{markov-ai/gaming-500-hours} dataset, and we sample computer-use and video game examples at a ratio of 1:3. Each example contains up to 20 screenshots for computer use and 40 frames for video games, with older steps given as text, and we compute the cross-entropy loss on every token of the target action.

\subsection{Router Architecture}
The router's input concatenates the visual tokens of $K+1=11$ observations, which gives $11\times2{,}048=22{,}528$ dimensions. The MLP has layers of size 22,528, 1,024, 256, and 1, with GELU activations and a dropout of 0.1, for a total of 23.3M parameters. Requests with a single observation are kept in full.

\subsection{Initialization}
\label{app:initialization}
For the initialization labels (\S\ref{sec:objective}), a region counts as changed if its mean absolute pixel change from the previous observation is at least 10 out of 255. We train on windows of 11 observations and supervise only the last one, using 47,489 computer-use windows and 29,709 video game windows. We train the router with a binary cross-entropy loss for 558 steps, with a batch size of 1,024 and AdamW with a learning rate of $1.5\times10^{-3}$.

\subsection{Joint Training}
Table~\ref{tab:hyperparameters} lists the hyperparameters for joint training. The full and blind models each have their own AdamW optimizer, and the visual tokens of each example are computed once and shared by all three models. For the adaptive weight, we use a 10 times larger step size when $\bar{r}$ is below $r^\star$, so that $\lambda$ decreases quickly and the router resumes dropping tokens.

\begin{table}[h]
\centering
\caption{Hyperparameters for joint training.}
\label{tab:hyperparameters}
\small
\begin{tabular}{@{}ll@{}}
\toprule
\textbf{Hyperparameter} & \textbf{Value} \\
\midrule
Base model & Qwen3.5-2B \\
Vision encoder & Frozen \\
Precision & BF16 \\
Context length & 65,536 tokens \\
Batch size & 32 \\
Hardware & 8 H100 GPUs \\
Training steps & 2,000 \\
Optimizer & AdamW \\
Model learning rate & $10^{-5}$, constant \\
Router learning rate & $5\times10^{-5}$ \\
Weight decay (model / router) & 0.01 / $10^{-4}$ \\
Gradient clipping (model) & 1.0 \\
Router parameters & 23.3M \\
History length $K$ & 10 \\
Anchor interval $B$ & 10 \\
Degradation target $r^\star$ & 0.1 \\
Step size $\alpha$ & 0.2 \\
Initial weight $\lambda_0$ & 0.5 \\
\bottomrule
\end{tabular}
\end{table}

\subsection{Baselines}
The dense, pixel-diff, and H.264 baselines are evaluated after 1,000 training steps, and the frozen router after 2,000.

\paragraph{Dense.} Images are resized to 100\%, 50\%, 25\%, and 12.5\% of their area, which gives 2,040, 1,008, 510, and 252 visual tokens per screenshot, and 220, 112, 55, and 28 per game frame.

\paragraph{Pixel-diff.} Pixel-diff uses the same threshold as the initialization labels (Appendix~\ref{app:initialization}).

\paragraph{H.264.} We encode the observations with libx264 at a quantization parameter of 0, with a keyframe every 10 frames to match the anchors. A region counts as changed if it has a nonzero motion vector or a nonzero residual, and keyframes are kept in full.

\paragraph{Frozen router.} The model is trained with the cross-entropy loss alone, without the reference models or the adaptive weight.

\paragraph{VTR at lower resolution.} Observations are downscaled by a linear factor of $1/\sqrt{2}$ or $1/2$ before encoding, with the same trained router and anchors.

\subsection{Evaluation}
\paragraph{OSWorld.} We evaluate on the 295 of the 369 tasks that do not require internet access, with up to 100 steps per task, and average the results over three seeds.

\paragraph{CUA-World.} We evaluate on 512 tasks, with up to 200 steps per task, and report the average score from its checklist verifier with Gemini 3 Flash.

\paragraph{Video games.} The held-out data contains about 90,000 actions from 36 games, consisting of 30,000 clicks, 26,880 mouse moves, 3,120 drags, 30,000 key presses, and 113 scrolls. A click counts as correct if the predicted action is a click with the same button and phase, press or release, as the target.

\paragraph{Cost and prefill time.} We compute the inference cost per task at prices of \$0.017, \$0.015 and \$0.025 per million input, cached input and output tokens, with prompt caching, and the prefill time as the time to prefill one request of the mean input length in vLLM on one NVIDIA L40S GPU.

\section{Additional Results}
\label{app:results}

\begin{table}[t]
\centering
\caption{Results on OSWorld and CUA-World for all methods. Tokens are visual tokens per image. Cost and prefill time are reductions relative to the dense model at full resolution.}
\label{tab:desktop-full}
\small
\setlength{\tabcolsep}{4pt}
\begin{tabular}{@{}lrrrrrrrr@{}}
\toprule
& \multicolumn{4}{c}{\textbf{OSWorld}} & \multicolumn{4}{c}{\textbf{CUA-World}} \\
\cmidrule(lr){2-5}\cmidrule(l){6-9}
\textbf{Method} & \textbf{Tokens} & \textbf{Score} & \textbf{Cost} & \textbf{Prefill} & \textbf{Tokens} & \textbf{Avg.\ score} & \textbf{Cost} & \textbf{Prefill} \\
\midrule
Dense, 100\% area & 2040.0 & 18.56 & 1.0$\times$ & 1.0$\times$ & 1946.2 & 16.61 & 1.0$\times$ & 1.0$\times$ \\
Dense, 50\% area & 1008.0 & 15.19 & 1.9$\times$ & 1.9$\times$ & 966.5 & 15.33 & 1.6$\times$ & 1.8$\times$ \\
Dense, 25\% area & 510.0 & 13.39 & 1.8$\times$ & 2.8$\times$ & 510.0 & 11.75 & 1.8$\times$ & 2.5$\times$ \\
Dense, 12.5\% area & 252.0 & 6.17 & 2.9$\times$ & 4.6$\times$ & 252.0 & 8.98 & 3.6$\times$ & 3.9$\times$ \\
\midrule
H.264 & 1386.5 & 9.44 & 1.3$\times$ & 1.4$\times$ & 1367.0 & 12.43 & 1.7$\times$ & 1.5$\times$ \\
Pixel-diff & 485.9 & 5.25 & 2.4$\times$ & 3.2$\times$ & 406.8 & 9.33 & 2.7$\times$ & 3.1$\times$ \\
Frozen router & 421.5 & 4.91 & 3.6$\times$ & 3.8$\times$ & N/A & N/A & N/A & N/A \\
\midrule
VTR (ours) & 807.9 & 18.10 & 2.4$\times$ & 2.3$\times$ & 674.3 & 16.34 & 1.8$\times$ & 2.2$\times$ \\
VTR sliding (ours) & 712.6 & 18.10 & 2.4$\times$ & 2.5$\times$ & 595.2 & 15.77 & 1.8$\times$ & 2.4$\times$ \\
VTR, 1/2 area (ours) & 391.0 & 13.73 & 2.8$\times$ & 3.6$\times$ & 328.8 & 12.92 & 2.2$\times$ & 3.2$\times$ \\
VTR, 1/4 area (ours) & 171.8 & 5.76 & 2.8$\times$ & 5.0$\times$ & 148.1 & 9.25 & 2.5$\times$ & 4.0$\times$ \\
\bottomrule
\end{tabular}
\end{table}

\begin{table}[t]
\centering
\caption{Difference between VTR and the dense model at full resolution, with a paired $t$-test over the tasks graded for both methods.}
\label{tab:significance}
\small
\setlength{\tabcolsep}{6pt}
\begin{tabular}{@{}llrcr@{}}
\toprule
\textbf{Benchmark} & \textbf{Metric} & \textbf{Difference} & \textbf{95\% CI} & \textbf{$p$-value} \\
\midrule
OSWorld & Score & $-0.46$ & $[-2.88,\ 1.94]$ & 0.71 \\
OSWorld & Pass rate (\%) & $-0.34$ & $[-2.71,\ 2.03]$ & 0.77 \\
CUA-World & Average score & $-0.52$ & $[-2.78,\ 1.74]$ & 0.66 \\
CUA-World & Pass rate (\%) & $+0.20$ & N/A & 0.87 \\
\bottomrule
\end{tabular}
\end{table}

\begin{table}[t]
\centering
\caption{Action prediction accuracy (\%) in video games for all methods, by action type. Reduction is relative to the dense model at full resolution.}
\label{tab:game-actions-full}
\small
\setlength{\tabcolsep}{5pt}
\begin{tabular}{@{}lrrrrrr@{}}
\toprule
\textbf{Method} & \textbf{Tokens} & \textbf{Reduction} & \textbf{Click} & \textbf{Move} & \textbf{Drag} & \textbf{Keyboard} \\
\midrule
Dense, 100\% area & 220.0 & 1.0$\times$ & 4.74 & 23.33 & 14.90 & 4.98 \\
Dense, 50\% area & 112.0 & 2.0$\times$ & 3.82 & 22.28 & 16.28 & 5.23 \\
Dense, 25\% area & 55.0 & 4.0$\times$ & 3.63 & 22.91 & 15.96 & 4.76 \\
Dense, 12.5\% area & 28.0 & 7.9$\times$ & 3.44 & 23.39 & 16.15 & 5.04 \\
\midrule
H.264 & 150.3 & 1.5$\times$ & 2.54 & 22.68 & 15.13 & 5.42 \\
Pixel-diff & 67.3 & 3.3$\times$ & 4.59 & 24.48 & 15.58 & 5.86 \\
Frozen router & 51.6 & 4.3$\times$ & 4.14 & 28.79 & 19.84 & 5.92 \\
\midrule
VTR (ours) & 38.2 & 5.8$\times$ & 4.90 & 24.94 & 16.86 & 5.37 \\
VTR sliding (ours) & 23.2 & 9.5$\times$ & 4.34 & 25.41 & 17.05 & 5.41 \\
VTR, 1/2 area (ours) & 16.4 & 13.4$\times$ & 4.54 & 24.19 & 16.38 & 4.84 \\
VTR, 1/4 area (ours) & 9.5 & 23.2$\times$ & 4.31 & 23.90 & 16.19 & 4.76 \\
\bottomrule
\end{tabular}
\end{table}

Table~\ref{tab:significance} tests whether VTR performs differently from the dense model at full resolution on OSWorld and CUA-World. On both benchmarks, the difference is small and not statistically significant.

\subsection{Qualitative Examples}
\label{app:qualitative}

\begin{figure}[h]
\centering
\setlength{\tabcolsep}{1.5pt}
\renewcommand{\arraystretch}{0.6}
\begin{tabular}{@{}cc@{\hspace{6pt}}cc@{}}
\multicolumn{2}{c}{\footnotesize GTA V: the camera follows the motorcycle} & \multicolumn{2}{c}{\footnotesize Candy Crush: the player moves one candy} \\[2pt]
\includegraphics[width=0.238\linewidth]{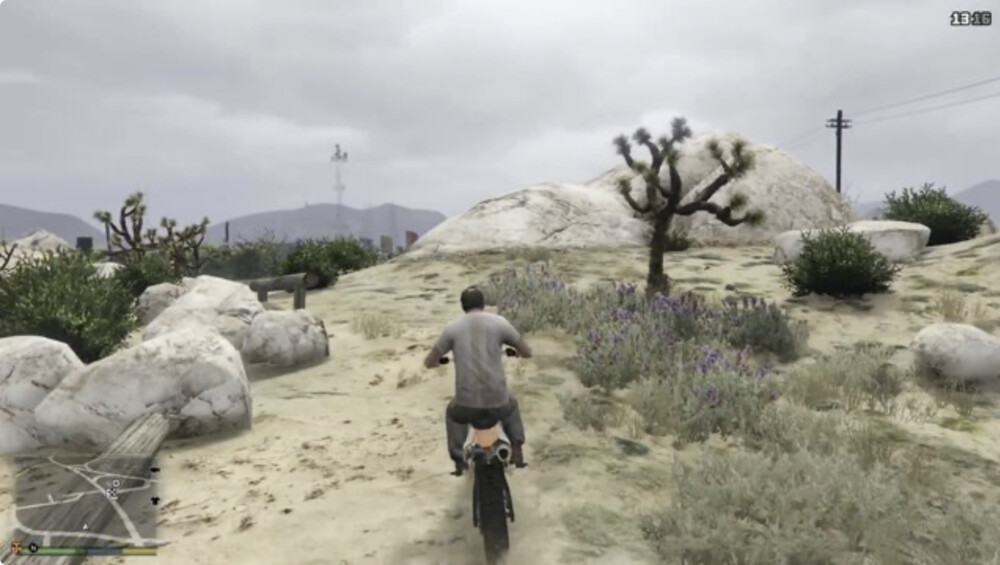} & \includegraphics[width=0.238\linewidth]{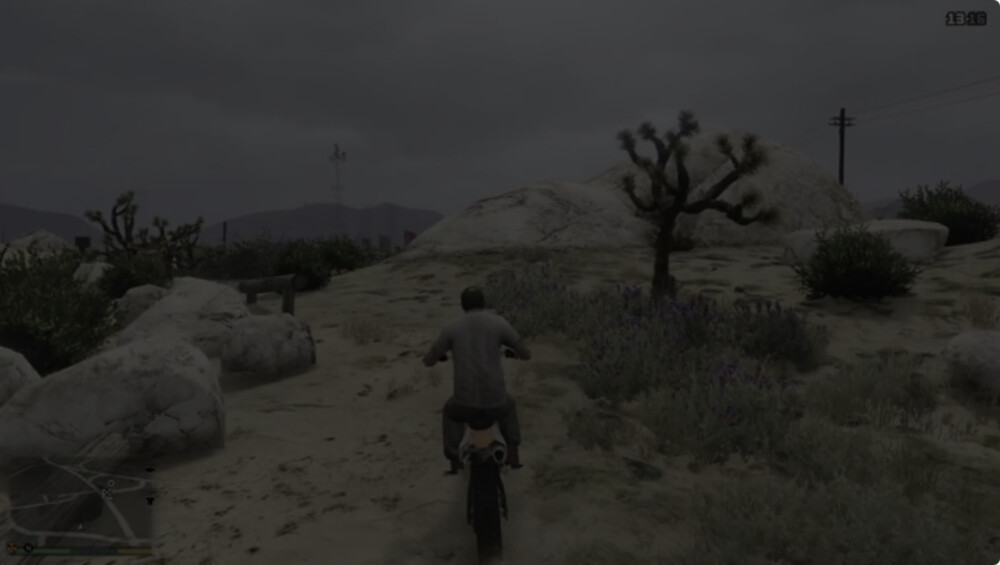} & \includegraphics[width=0.238\linewidth]{figures/qualitative/candy_prev.jpg} & \includegraphics[width=0.238\linewidth]{figures/qualitative/candy_kept.jpg} \\[6pt]
\multicolumn{2}{c}{\footnotesize Candy Crush: new candies fall into the board} & \multicolumn{2}{c}{\footnotesize Minecraft: a fight with another player} \\[2pt]
\includegraphics[width=0.238\linewidth]{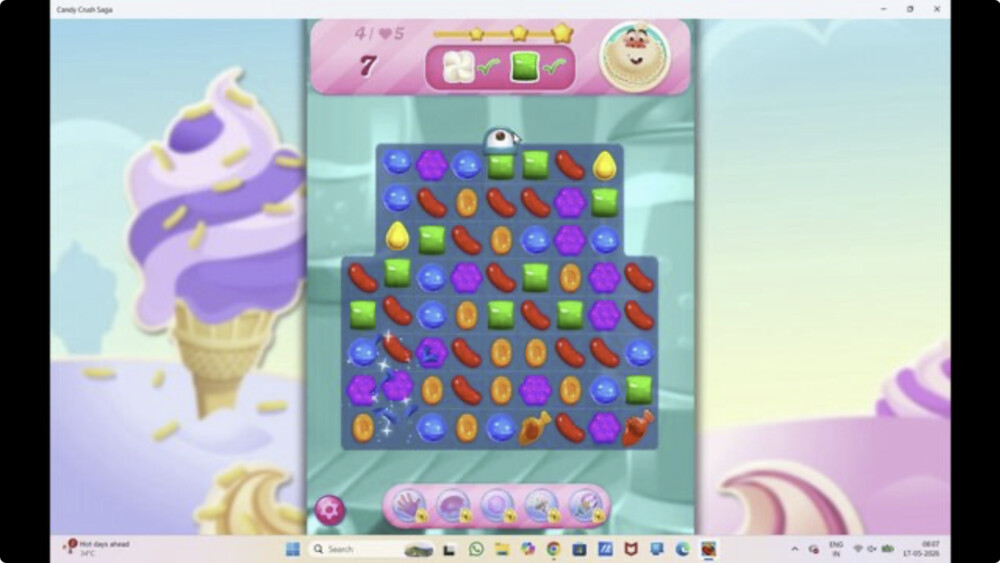} & \includegraphics[width=0.238\linewidth]{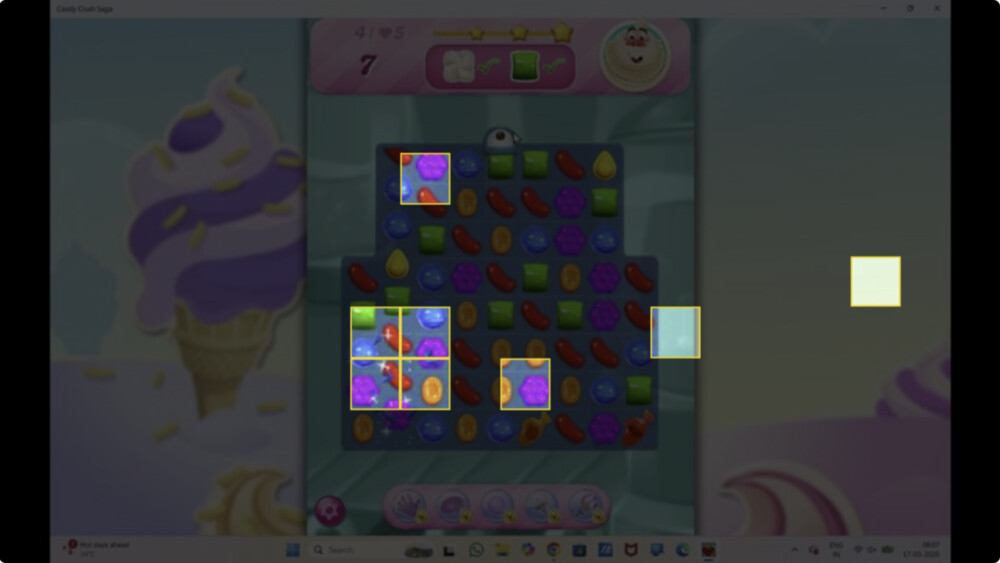} & \includegraphics[width=0.238\linewidth]{figures/qualitative/minecraft_prev.jpg} & \includegraphics[width=0.238\linewidth]{figures/qualitative/minecraft_kept.jpg} \\[6pt]
\multicolumn{2}{c}{\footnotesize Dota 2: a fight on an otherwise static map} & \multicolumn{2}{c}{\footnotesize Gpredict: the agent types a satellite name} \\[2pt]
\includegraphics[width=0.238\linewidth]{figures/qualitative/dota_prev.jpg} & \includegraphics[width=0.238\linewidth]{figures/qualitative/dota_kept.jpg} & \includegraphics[width=0.238\linewidth]{figures/qualitative/gpredict_prev.jpg} & \includegraphics[width=0.238\linewidth]{figures/qualitative/gpredict_kept.jpg} \\[6pt]
\multicolumn{2}{c}{\footnotesize LibreCAD: the agent opens a menu} & \multicolumn{2}{c}{\footnotesize PsychoPy: the agent names a new routine} \\[2pt]
\includegraphics[width=0.238\linewidth]{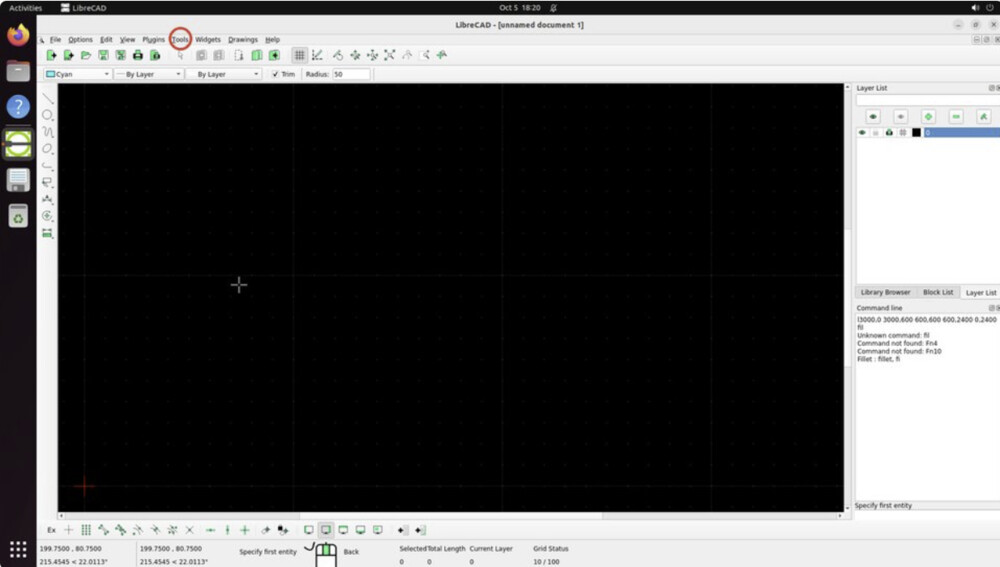} & \includegraphics[width=0.238\linewidth]{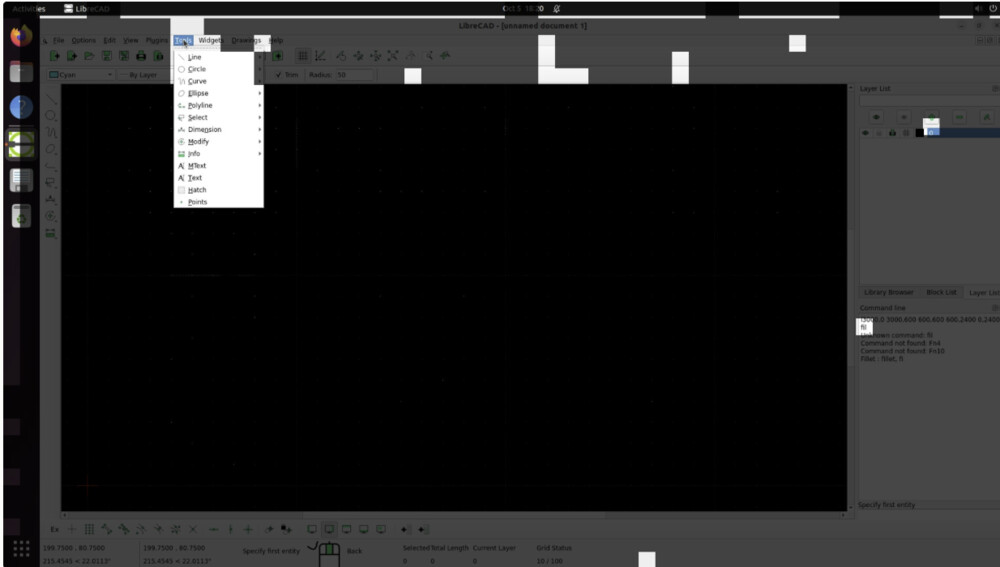} & \includegraphics[width=0.238\linewidth]{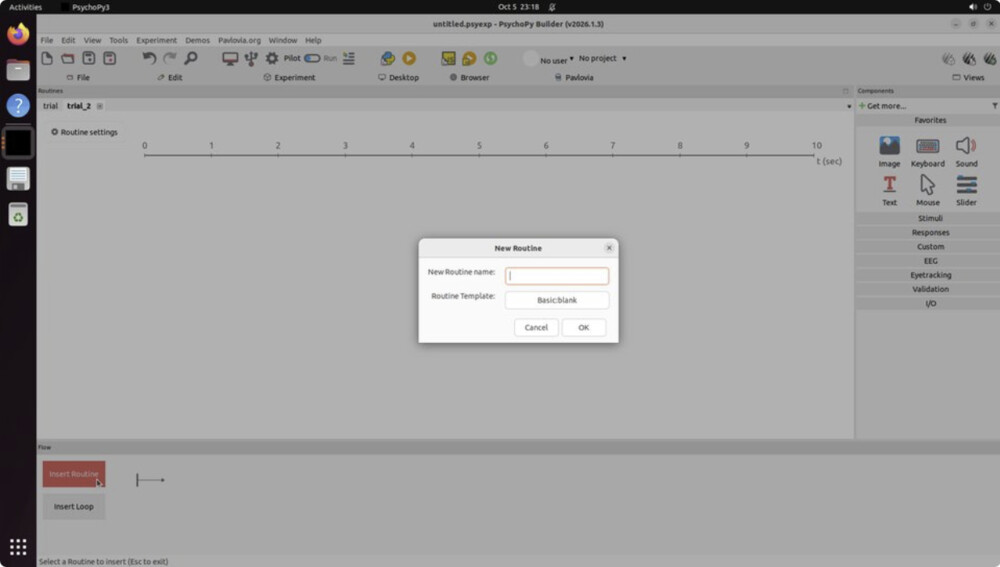} & \includegraphics[width=0.238\linewidth]{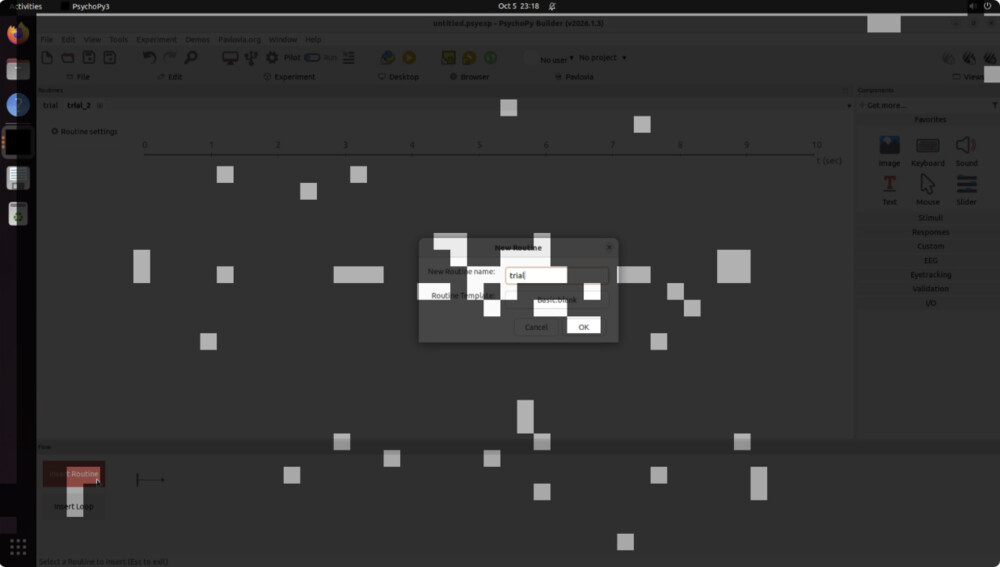} \\
\end{tabular}
\caption{\textbf{Further examples of the regions that VTR keeps.} For each example, the previous observation (left) and the current observation with the regions that VTR drops dimmed (right), in video games and desktop applications.}
\label{fig:qualitative-all}
\end{figure}

Figure~\ref{fig:qualitative-all} shows further examples of the regions that VTR keeps, in video games and desktop applications.

\section{Extended Related Work}
\label{app:related}

This appendix expands on the related work in \S\ref{sec:related}.

\paragraph{Visual token reduction.} Vision-language models represent each image with hundreds to thousands of visual tokens, and a large body of work reduces this number to speed up inference. Training-free methods score tokens by attention, either inside the language model, as FastV~\citep{chen2024fastv} does after its second layer, or in the vision encoder, as VisionZip~\citep{yang2025visionzip} does, while SparseVLM~\citep{zhang2025sparsevlm} scores visual tokens by the attention they receive from the relevant text tokens. Token merging combines similar tokens instead of dropping them, first in vision transformers~\citep{bolya2023tome} and then in vision-language models, where LLaVA-PruMerge~\citep{shang2025prumerge} also chooses the number of kept tokens for each image with an outlier rule on attention scores. For video, methods exploit redundancy across frames: LongVU~\citep{shen2025longvu} drops near-duplicate frames and reduces tokens based on their similarity across frames, EVS~\citep{bagrov2025evs} removes patches that stay the same across consecutive frames while keeping the positions of the remaining tokens, and run-length tokenization~\citep{choudhury2024rlt} replaces patches that repeat over time with a single token. Codec-based methods instead read the motion vectors and residuals that video codecs already compute: OneVision-Encoder~\citep{tang2026ovencoder} selects the patches with the strongest motion and residuals across frames, up to a fixed token budget, and trains its vision encoder on these sparse inputs, while CoPE-VideoLM~\citep{sarkar2026cope} encodes only keyframes with the image encoder and represents the other frames with a few tokens built from motion vectors and residuals. Learned methods train a module that predicts which tokens to keep together with the model, starting with DynamicViT~\citep{rao2021dynamicvit} for image classification and extending to vision-language models with Dynamic-LLaVA~\citep{huang2025dynamicllava} and ATP-LLaVA~\citep{ye2025atpllava}, while VideoLLM-MoD~\citep{wu2024videollmmod} learns which video tokens skip the computation of each layer. However, in all of these methods, the number of kept tokens is set by hand: training-free methods use a fixed ratio or threshold, codec-based methods use a fixed token budget or keyframe density, DynamicViT and Dynamic-LLaVA add a loss that pulls the kept fraction toward a pre-defined keep rate, ATP-LLaVA pulls the average number of kept tokens toward a target count, and VideoLLM-MoD skips a fixed fraction of tokens. Moreover, apart from the video methods, they decide for each image independently. SCORE~\citep{wang2026score} comes closest to our work. A small MLP reads each video token together with its change since the previous frame and outputs a keep probability, and the MLP is trained with reinforcement learning on sampled keep masks. However, SCORE keeps the vision encoder and the language model frozen, looks only one frame back, and keeps a fixed fraction of tokens at inference by selecting the highest-scoring ones. In contrast, the router in VTR reads each image region across several previous screenshots and is trained jointly with the language model, so the model learns to act from the kept tokens, and the number of kept tokens is not set in advance but chosen so that the model does not lose accuracy.

\paragraph{Efficient computer-use agents.} As computer-use agents improve on benchmarks such as OSWorld~\citep{xie2024osworldbenchmarkingmultimodalagents}, recent work has turned to their speed and cost. OSWorld-Human~\citep{abhyankar2026osworldhumanbenchmarkingefficiencycomputeruse} finds that agents take tens of minutes on tasks that humans complete in a few minutes, with later steps taking up to three times longer than early ones, and cua-speedrun~\citep{aggarwal2026cuaspeedrun} standardizes the infrastructure for measuring the speed and cost of agents across benchmarks. In real-time domains, VideoGameBench~\citep{zhang2025videogamebench} identifies inference latency as a major limitation of frontier models in video games, and in Gym-Anything-RT~\citep{aggarwal2026gymanythingrt}, the pass rate of a frontier agent rises from 9.0\% to 75.0\% when the environment is paused while the agent decides. To limit the cost of screenshots, agents commonly keep only the last few screenshots in context and represent earlier steps as text~\citep{qin2025uitars,wang2025opencua}, and Anthropic recommends reducing the resolution of screenshots and keeping only the most recent ones~\citep{anthropic2026computeruse}. Methods designed for GUI agents reduce tokens more selectively. ShowUI~\citep{lin2025showui} groups patches of similar color into connected components and skips part of the patches in each component, GUIPruner~\citep{xu2026guipruner} lowers the resolution of older screenshots and prunes the current one with rules that preserve its layout, GUI-KV~\citep{huang2026guikv} compresses the key-value cache by scoring how redundant older screenshots are given the current one, and SimpAgent~\citep{chen2025simpagent} trains the model to drop all history screenshots after a fixed layer. STP-GUI~\citep{wang2026stpgui} learns a token selector with reinforcement learning, but keeps the model frozen and keeps a fixed fraction of tokens. However, none of these methods learns which tokens to keep together with the model that predicts actions. ReVision~\citep{abaskohi2026revision} comes closest to our work. A three-layer MLP compares each patch with the same patch in the previous screenshot and drops unchanged patches, the first screenshot in the window is kept in full, and kept patches retain their original positions. However, ReVision trains its MLP on labels obtained by matching regions from a UI parser across screenshots, and only then fine-tunes the model on the filtered trajectories, so the MLP detects unchanged content rather than learning what the agent needs for its next action. In contrast, VTR starts its router from similar labels, which combine changed regions with segmented objects (\S\ref{sec:objective}), and then trains the router jointly with the model to preserve action prediction, on desktop tasks and video games.

\paragraph{Learned routing and constrained optimization.} Conditional computation trains a router to decide which parts of a network process each input. Sparsely gated mixture-of-experts models route each token to its top-scoring experts~\citep{shazeer2017moe,fedus2022switch}, expert choice routing lets each expert pick a fixed number of tokens~\citep{zhou2022expertchoice}, and Mixture-of-Depths~\citep{raposo2024mod} applies the same idea across layers, sending a fixed number of tokens through each block while the rest skip it. In these models, the router's weight scales the output, which gives the router a gradient despite the discrete choice, while other methods train discrete gates with the straight-through estimator~\citep{bengio2013ste} or with continuous relaxations such as Gumbel-softmax~\citep{jang2017gumbel,maddison2017concrete}. Adaptive computation methods let the amount of computation vary across inputs, by learning when to stop computing under a penalty on the amount of computation~\citep{graves2016act,yin2022avit}, or by activating a variable number of experts~\citep{wang2025remoe,guo2025dynmoe}, where ReMoE~\citep{wang2025remoe} adapts the weight of its sparsity penalty to reach a target average sparsity. Separately, constrained optimization replaces hand-tuned loss weights with Lagrange multipliers that are adapted during training to meet a target, as in constrained VAEs~\citep{rezende2018geco}, constrained reinforcement learning~\citep{altman1999cmdp,tessler2019rcpo,stooke2020pid}, and pruning to a target sparsity~\citep{wang2020structured,xia2022cofi,gallegoposada2022controlled}. However, methods that reduce computation fix its amount, set a target for it, or weight it with a hand-tuned penalty, so they do not ensure that the model keeps its accuracy. CALM~\citep{schuster2022calm} comes closest to our work, letting each token exit the language model early and calibrating the exit thresholds after training so that, with high probability, the outputs stay close to those of the full model. In contrast, VTR minimizes the number of kept tokens subject to the model not losing accuracy, measured against reference models with and without the full history, and enforces this constraint during training by dual ascent. Like mixture-of-experts models, VTR scales each kept token by its keep probability, but it removes dropped tokens from the input entirely, so they cost no computation in any layer.

\end{document}